%% file: main.tex
\documentclass[lettersize,journal]{IEEEtran}
\usepackage{pifont}
\usepackage{amsmath,amsfonts}
\usepackage{algorithmic}
\usepackage{array}
\usepackage[caption=false,font=normalsize,labelfont=sf,textfont=sf]{subfig}
\usepackage{textcomp}
\usepackage{stfloats}
\usepackage{url}
\usepackage{verbatim}
\usepackage{graphicx}

\usepackage{dsfont}
\usepackage{multirow}
\usepackage{multicol}
\usepackage{booktabs}
\usepackage{color}
\usepackage{xcolor}
\usepackage{float}

\usepackage[framemethod=tikz]{mdframed}
\usepackage{soul}
\soulregister{\cite}7 
\soulregister{\citep}7 
\soulregister{\citet}7 
\soulregister{\ref}7 
\soulregister{\textbf}7
\usepackage{cancel}

\usepackage{pifont}
\usepackage{color, colortbl}
\definecolor{Gray}{gray}{0.9}
\definecolor{darkergreen}{RGB}{21, 152, 56}
\definecolor{red2}{RGB}{252, 54, 65}

\def\BibTeX{{\rm B\kern-.05em{\sc i\kern-.025em b}\kern-.08em
    T\kern-.1667em\lower.7ex\hbox{E}\kern-.125emX}}
\usepackage{balance}

\begin{document}
\title{SSP-SAM: SAM with Semantic-Spatial Prompt for Referring Expression Segmentation}
\author{Wei Tang, Xuejing Liu, Yanpeng Sun, and Zechao Li, \textit{Senior Member, IEEE}
\IEEEcompsocitemizethanks{
\IEEEcompsocthanksitem W. Tang and Z. Li are with School of Computer Science and Engineering, Nanjing University of Science and Technology, No. 200 Xiaolingwei Road, Nanjing 210094, China. E-mail: \{weitang, zechao.li\}@njust.edu.cn; 
\IEEEcompsocthanksitem Y. Sun is with NExT++ Lab, School of Computing, National University of Singapore, Singapore 119077, Singapore. E-mail: yanpeng\_{sun}@nus.edu.sg;   
\IEEEcompsocthanksitem X. Liu is with Institute of Computing Technology, Chinese Academy of Sciences, Beijing 100190, China. E-mail: xuejing.liu@vipl.ict.ac.cn.

(Corresponding authors: Zechao Li and Xuejing Liu)
  }
}


\makeatletter
\def\ps@IEEEtitlepagestyle{
  \def\@oddfoot{\mycopyrightnotice}
  \def\@evenfoot{}
}
\def\mycopyrightnotice{
  {\footnotesize
  \begin{minipage}{\textwidth}
  \centering
  Copyright~\copyright~2025 IEEE. Personal use of this material is permitted. However, permission to use this  \\ 
  material for any other purposes must be obtained from the IEEE by sending a request to pubs-permissions@ieee.org.
  \end{minipage}
  }
}

\IEEEpubidadjcol
\maketitle

\begin{abstract}
The Segment Anything Model (SAM) excels at general image segmentation but has limited ability to understand natural language, which restricts its direct application in Referring Expression Segmentation (RES). Toward this end, we propose SSP-SAM, a framework that fully utilizes SAM’s segmentation capabilities by integrating a Semantic-Spatial Prompt (SSP) encoder. Specifically, we incorporate both visual and linguistic attention adapters into the SSP encoder, which highlight salient objects within the visual features and discriminative phrases within the linguistic features. This design enhances the referent representation for the prompt generator, resulting in high-quality SSPs that enable SAM to generate precise masks guided by language. Although not specifically designed for Generalized RES (GRES), where the referent may correspond to zero, one, or multiple objects, SSP-SAM naturally supports this more flexible setting without additional modifications. Extensive experiments on widely used RES and GRES benchmarks confirm the superiority of our method. Notably, our approach generates segmentation masks of high quality, achieving strong precision even at strict thresholds such as Pr@0.9. Further evaluation on the PhraseCut dataset demonstrates improved performance in open-vocabulary scenarios compared to existing state-of-the-art RES methods. The code and checkpoints are available at: https://github.com/WayneTomas/SSP-SAM.
\end{abstract}


\section{Introduction}
\input{section/intro}

\section{Related Work}
\label{relat_RIS}
\input{section/related_work}

\section{Method}
\begin{figure*}[t]
    \centering
    \setlength{\belowcaptionskip}{-5pt} 
    \includegraphics[width=1\linewidth]{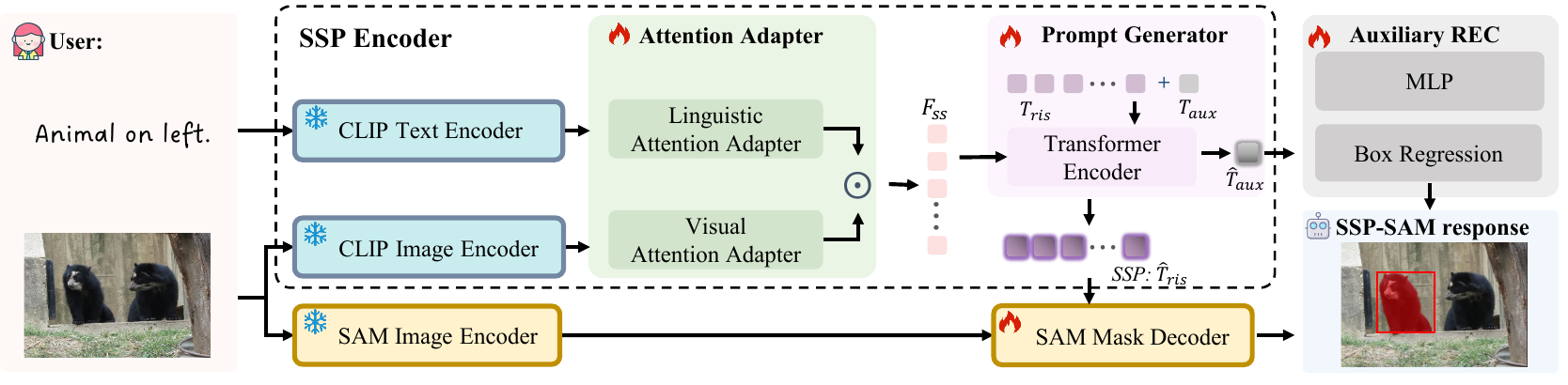}
    \caption{An illustration of SSP-SAM, where SAM is equipped with Semantic-Spatial Prompt (SSP) encoder for RES. The multi-modal features of inputs are extracted by CLIP, and their attention scores are adapted to generate Semantic-Spatial referent features. These features, along with the special tokens, are processed by a prompt generator to identify key information about the referent for the segmentation process. Additionally, an auxiliary REC task is utilized to improve RES performance.}
    \label{fig:framework}
\end{figure*}

\begin{figure}[t]
    \centering
    \includegraphics[width=0.9\linewidth]{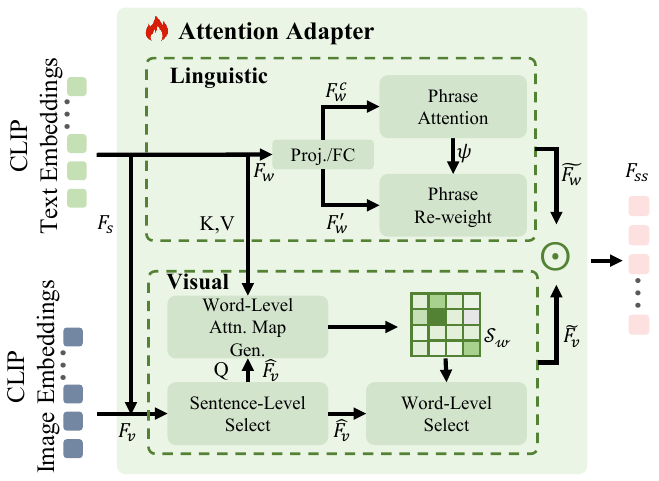}
    \caption{An illustration of the attention adapter. It consists of a visual attention adapter and a linguistic attention adapter. The images and texts embedding are extracted from CLIP and then feed to the visual and linguistic attention adapter to obtain the enhanced referent features.}
    \label{fig:attn_adapter}
    \vspace{-4mm}
\end{figure}
\subsection{Overview}
Given an image $\mathcal{I}$ and a language query $\mathcal{Q}$, RES aims to predict a binary mask $\mathcal{M}$ for the referent in the query. We build on SAM’s strengths with a unified one-stage approach, where the SSP encoder serves as a core module. It consists of two main components. The first is the attention adapter module, which has two sub-components: the \textit{visual attention adapter}, highlighting salient objects explicitly mentioned in the expression (mostly nouns), and the \textit{linguistic attention adapter}, assigning higher weights to phrases that capture abstract relations or attributes to help distinguish the referent from other objects. Their outputs are combined via element-wise product, with the linguistic features acting as an attention mask that selectively enhances the relevant visual features. The second main component is the prompt generator, which fuses these enhanced features with semantic and spatial cues to produce the final prompts for SAM. Furthermore, to enhance the quality of the prompts, we introduce an auxiliary task.

\subsection{Attention Adapter}

The proposed attention adapter consists of two sub-modules, one focusing on visual attention and the other on linguistic attention. The details are shown in Fig.~\ref{fig:attn_adapter}.

In the \textit{visual attention adapter}, to leverage CLIP’s image-level knowledge, we first compute sentence-level similarities to coarsely filter visual features relevant to the given text.  
Specifically, let \( F_v \in \mathbb{R}^{B \times N \times D} \) denote the visual features extracted from the CLIP visual encoder, 
where \( B \) is the batch size, \( N \) is the number of patch embeddings, and \( D \) is the feature dimension.  
Let \( F_s \in \mathbb{R}^{B \times D} \) be the sentence-level embedding from the CLIP text encoder. The sentence-level attention map is computed as:
\begin{equation}
\label{eq_sim_map}
    \mathcal{S}_c = \mathrm{norm}(F_v) \cdot \mathrm{norm}(F_s)^\top ,
\end{equation}
where \( \mathrm{norm}(\cdot) \) denotes L2 normalization.  
{Conceptually, for each batch element $b$, the sentence embedding $F_s[b,:]$ is broadcasted to match the patch dimension $N$ of $F_v[b,:,:]$, so that the similarity with each visual token can be computed, resulting in $\mathcal{S}_c \in \mathbb{R}^{B \times N \times 1}$.}
We then select the sentence-level visual features through an element-wise product:
\begin{equation}
\label{eq_select_feat}
    \hat{F_v} = F_v \odot \mathcal{S}_c.
\end{equation}

However, locating the language-conditioned referent requires more detailed information. Therefore, we propose employing word-level visual attention adaptation to further refine the selected sentence-level visual features. Let \( F_w \in \mathbb{R}^{B \times L \times D} \) denote the word-level features extracted from the CLIP text encoder, where \( L \) is the number of tokens (i.e., 77 for CLIP-ViT-B/16-224 and CLIP-ViT-L/14-336). The word-level attention map \( \mathcal{S}_w \in \mathbb{R}^{B \times N \times D'} \) is defined as:
\begin{equation}
    \begin{aligned}
        \mathcal{S}_w = S(\hat{F_v}, \mathrm{CrossAttn}(\hat{F_v}, F_w)),
    \end{aligned}
\end{equation}
where \(\mathrm{CrossAttn}(\hat{F_v}, F_w)\) denotes a cross-attention operation that treats the selected visual features \(\hat{F_v}\) as queries and the word features \(F_w\) as keys and values, thereby capturing word-level semantics. The attention map \(\mathcal{S}_w\) is then generated by computing the similarity between the selected sentence-level visual features \(\hat{F_v}\) and the aggregated semantic features using a similarity function \(S(\cdot)\).

The more fine-grained visual features are selected using word-level features, which highlight salient objects with high attention scores. Following the successful practices of VLTVG \cite{vltvg} and TransCP \cite{transcp}, we also adopt a Gaussian-like mapping to further enhance the discrimination between the salient objects and the context:
 \begin{equation}
    \begin{aligned}
        \widetilde{F_v} = {\mathrm{FC}(\hat{F_v}}) \odot (\alpha \cdot \exp(-\frac{(1-\mathcal{S}_w)^2}{2\delta^2})) \in \mathbb{R}^{B \times N \times D'},
     \end{aligned}
     \label{word-level-select}
\end{equation}
where $\alpha$ and $\delta$ are two learnable parameters that are empirically initialized to $1$ and $0.5$. $\mathrm{FC}(\cdot)$ denotes a fully-connected layer.
By leveraging Eq.~\ref{word-level-select}, the salient objects within the selected visual features are assigned higher attention scores, effectively integrating spatial information into the semantic representation of the referent. 
{The corresponding visualization can be found in Fig.~\ref{fig:visualization_attn_visual}, where salient objects with high attention values are highlighted in red. Overall, in the \textit{visual attention adapter}, we perform feature selection by leveraging both the sentence-level and word-level attention maps.}

To further improve the spatial information for the referent, we propose a \textit{linguistic attention adapter} that focuses on capturing more discriminative word features related to the referent, including attributes and relationships. 
Given the original CLIP word embeddings \(F_w \in \mathbb{R}^{B \times L \times D}\), we apply a projection module \(\mathrm{Proj}(\cdot)\), comprising a fully-connected layer and a single-layer bi-GRU \cite{gru}, to obtain context-aware features \(F_w^c\):
\begin{equation}
    F_w^{c} = \mathrm{Proj}(F_w).
\end{equation}
Simultaneously, we apply a fully-connected layer to \(F_w\) to obtain projected word features:
\begin{equation}
    F_w' = \mathrm{FC}(F_w).
\end{equation}
Then, phrase attention weights \(\psi\) are computed over the context-aware features \(F_w^c\):
\begin{equation}
    \psi = \mathrm{softmax}(\mathrm{FC}(F_w^{c})) \in \mathbb{R}^{B \times 1 \times L},
\end{equation}
which are used to weight the projected features \(F_w'\) and produce the final re-weighted phrase feature:
\begin{equation}
    \widetilde{F_w} = \sum_{t=1}^{L} \psi_t \cdot F_w'^{(t)} \in \mathbb{R}^{B \times 1 \times D'}.
\end{equation}


The corresponding visualization can be found in Fig.~\ref{fig:visualization_attn_linguistic}, where the discriminative phrases within the referring expressions receive higher attention scores (e.g. \textit{``left"}, \textit{``larger"}, etc.). Overall, phrase attention in the \textit{linguistic attention adapter} re-weights word features, emphasizing phrases like attributes and spatial/relational cues that distinguish the referent from its context.

Finally, the enhanced multi-modal features that integrate both semantic and spatial cues are derived via gated fusion:
\begin{equation}
    F_{SS} = \tanh(\widetilde{F_v}) \odot \tanh(\widetilde{F_w}) \in \mathbb{R}^{B \times N \times D'},
\end{equation}
where \(\widetilde{F_v}\) are the selected visual features. 
The re-weighted phrase features \(\widetilde{F_w}\) serve as a soft semantic-spatial mask via element-wise interaction, enabling the visual features to be modulated by referent-discriminative linguistic cues.

\subsection{Prompt Generator}
As illustrated in Fig.~\ref{fig:framework}, we introduce a set of learnable special tokens $T_{res} = [t_{res}^0, t_{res}^1, ..., t_{res}^n]$ with a dimensionality of 256. These tokens are designed to extract referent-specific features from the Semantic-Spatial referent features $F_{SS}$, thereby guiding SAM in its segmentation endeavors. Specifically, we initialize these special tokens randomly and concatenate them with the previously obtained Semantic-Spatial referent features. Subsequently, these tokens are fed into a transformer encoder, which facilitates both intra- and inter-modal interactions.
 \begin{equation}
    \begin{aligned}
        \hat{T}_{res} = E_{r}([T_{res}, \text{MLP}(F_{SS})]),
     \end{aligned}
\end{equation}
where $E_{r}$ is the transformer encoder and $\text{MLP}(\cdot)$ is a 3-layer multilayer perceptron. Additionally, learnable position embeddings are incorporated into this encoder to retain the positional information of the tokens. 
The final state of the special tokens $T_{res}$, denoted as $\hat{T_{res}}$, serves as the Semantic-Spatial Prompts. These prompts are combined with the image features extracted by SAM’s image encoder and jointly fed into the SAM mask decoder to guide segmentation. This design maximizes the inherent segmentation capability of SAM by guiding it with discriminative, language-derived semantic and spatial cues.

To better leverage existing large-scale image-text paired data with bounding box annotations, we extend our framework with an auxiliary referring expression comprehension (REC) task, inspired by prior works~\cite{reftr, seqtr, polyformer, vg-law}. This extension is made possible by the flexible token-based design of our prompt generator, which allows the RES framework to be seamlessly adapted into a multi-task setup. Specifically, following the architectural paradigm of TransVG~\cite{transvg} and TransCP~\cite{transcp}, our framework allows for seamless extension by incorporating an additional special token $T_{aux}$ and a bounding box regression head. The final input to the prompt generator is thus reformulated as:
 \begin{equation}
    \begin{aligned}
        \hat{T}_{aux}, \hat{T}_{res} = E_{r}([T_{aux}, T_{res}, \text{MLP}(F_{SS})]).
     \end{aligned}
\end{equation}
In detail, the number of special tokens $T_{res}$ is empirically set to $128$, and more information can be found in \textbf{Section~\ref{sec:num_ris}}. The bounding box regression head consists of a 3-layer MLP that maps the auxiliary token $\hat{T}_{aux}$ to a 4-dimension sequence formatted as $(x, y, w, h)$. The auxiliary REC task not only provides better prompts, thereby improving the segmentation performance of SAM, but also allows for the full exploitation of box-annotated data for pre-training the SSP encoder, further enhancing the performance of the proposed framework (more details are shown in \textbf{Section~\ref{training_strategy}}).

\subsection{Training and Losses}
The proposed framework concisely bridges SAM and CLIP for the RES task, which can be optimized in an one-stage manner. During training, we incorporate the REC task as an auxiliary task to fully utilize bounding box annotations. 

Following \cite{seqtr, reftr, polyformer}, we first pre-train our model on a merged REC dataset for a pre-defined number of epochs, and then we fine-tune the proposed framework with both the RES and auxiliary REC tasks together on downstream datasets. Further details are provided in \textbf{Section~\ref{training_strategy}}.

Following recent research \cite{reftr, seqtr, polyformer, vltvg, transcp, transvg}, we utilize two types of losses to supervise the learning of the proposed prompt encoder. The first type consists of the Focal loss and the Dice loss, which are used to supervise the correspondence between the predicted segmentation mask and the ground-truth mask. Specifically, given the predicted binary segmentation mask $\hat{\mathcal{M}}$ and the ground-truth mask $\mathcal{M}$, the segmentation loss is defined as follows:
 \begin{equation}
    \begin{aligned}
        \mathcal{L}_{res} = \lambda_{Focal} \mathcal{L}_{Focal}(\mathcal{M}, \hat{\mathcal{M}}) + \lambda_{Dice} \mathcal{L}_{Dice}(\mathcal{M}, \hat{\mathcal{M}}).
     \end{aligned}
\end{equation}
The second type of loss is the combination of L1 loss and GIoU loss, which are used to guide the auxiliary task. Given the predicted bounding box $\hat{\mathcal{B}}$ and the ground-truth bounding box $\mathcal{B}$, the 
 \begin{equation}
    \begin{aligned}
        \mathcal{L}_{aux} = \lambda_{l1} \mathcal{L}_{l1}(\mathcal{B}, \hat{\mathcal{B}}) + \lambda_{GIoU} \mathcal{L}_{gIoU}(\mathcal{B}, \hat{\mathcal{B}}),
     \end{aligned}
\end{equation}
where L1 loss ensures numeric accuracy, and gIoU loss improves the bounding box localization by penalizing predictions that are too large or too small in comparison to the ground truth.
{Following \cite{reftr, vltvg, transcp}, $\lambda_{Focal}$, $\lambda_{Dice}$, $\lambda_{l1}$ and $\lambda_{GIoU}$ are set to $4$, $4$, $5$ and $2$ empirically.}
By combining the two types of losses, our framework seamlessly incorporates the auxiliary REC task to guide the Semantic-Spatial Prompt encoder, resulting in a significant enhancement of subsequent SAM segmentation performance. The total loss used for training is as follows:
 \begin{equation}
 \label{eq_loss_total}
    \begin{aligned}
       \mathcal{L}_{total} = \beta \mathcal{L}_{res} + \mathcal{L}_{aux}.
     \end{aligned}
\end{equation}
{$\beta$ is set to $0$ during pre-training since only auxiliary REC task is active, and switched to $1$ during fine-tuning to jointly optimize RES and REC.}

\section{Experiment}
\label{exp}

\subsection{Datasets}
\noindent\textbf{RefCOCO/+/g.}
RefCOCO, RefCOCO+, and RefCOCOg are classic RES benchmarks built on MS-COCO~\cite{mscoco}.
RefCOCO has 142k expressions for 50k objects (19,994 images) with testA (people) and testB (other objects).
RefCOCO+ removes location words (e.g., \textit{left}, \textit{right}) to emphasize appearance reasoning, including 141k expressions for 49k objects.
RefCOCOg contains longer, descriptive expressions (95k for 49k objects, 25k images), with two versions—google~\cite{refcocog_google} and umd~\cite{refcoco_umd}; we follow the latter.

{\noindent\textbf{ReferIt.}
ReferIt~\cite{referit} is another classic RES benchmark with 20k images and 120k expressions.
It features diverse scenes and ambiguous queries (e.g., \textit{any}, \textit{whole}), posing additional challenges beyond COCO-based datasets.}

{\noindent\textbf{gRefCOCO.}
To evaluate under the generalized RES (GRES)~\cite{gres} setting, we use gRefCOCO, which introduces no-target, single-target, and multi-target expressions (278k in total, 80k multi, 32k none) across 19,994 images, split into val/testA/testB.}

\noindent\textbf{PhraseCut.}
We also validate open-vocabulary segmentation on PhraseCut~\cite{phrasecut}, which contains 1,287 categories (vs. 80 in COCO) and 14,354 expressions over 2,545 test images.

\subsection{Implementation Details}
\label{implementation}
\noindent\textbf{Evaluation Metrics.}
Following prior works~\cite{gres, gsva}, we adopt generalized IoU (\textbf{gIoU}), cumulative IoU (\textbf{cIoU}), Precision@X \textbf{(Pr@X)}, and No-target accuracy (\textbf{N-Acc.}) as our evaluation metrics.
Specifically, \textbf{cIoU} computes the total intersection pixels over total union pixels across the dataset, while \textbf{Pr@X} indicates the percentage of samples with IoU exceeding a threshold $X$ (e.g., 0.5, 0.7, 0.9).
\textbf{gIoU} is the average of the per-instance IoUs, mitigating cIoU’s bias toward larger objects \cite{gres, phrasecut, lavt}. As such, we prioritize gIoU in our analysis. If not otherwise stated, we mean gIoU.

For \textbf{no-target} cases, we follow GRES~\cite{gres} which considers a predicted mask with fewer than 50 positive pixels as a valid empty prediction (IoU = 1). This metric measures the model’s accuracy in correctly identifying no-target samples.

\noindent \textbf{Training details.}
We report the performance of default SAM-H equiped with our encoder using two different CLIP versions, specifically CLIP-ViT-B/16-224 and CLIP-ViT-L/14-336, and the CLIP visual branch is the revised version introduced in \cite{li2023clip_surgery}. We refer to the configuration of the SSP encoder with the former backbone as SSP-SAM-224 and the latter as SSP-SAM-336. The number of learnable [RES] tokens is set to $128$ empirically, with more details provided in the ablation study of \textbf{Section~\ref{sec:num_ris}}. 

Following \cite{mcn, seqtr, reftr, polyformer}, we initially pre-train SSP encoder on a merged dataset for the auxiliary REC task. During this phase, the encoders and decoders of both the CLIP and SAM models are kept frozen. Following PolyFormer \cite{polyformer}, the pre-training epoch is empirically set to $20$. This merged pre-training dataset includes the REC training data from RefCOCO, RefCOCO+, RefCOCOg, ReferIt, Flickr30k Entities \cite{flickr30k}, and Visual Genome \cite{visual_genome}. During fine-tuning, our framework is tuned across $80$ epochs. 
We keep the SAM decoder frozen for the first $20$ epochs, while the remaining SAM components and the entire CLIP model remain frozen. Additionally, a batch size of $128$ is used for pre-training, while a batch size of $64$ is employed for fine-tuning. The entire training phase utilizes $8$ NVIDIA Tesla V100 GPUs. The initial learning rate is set to 1e-4, and a cosine decay strategy with a $5$-epoch warm-up is utilized. 

\textit{{Note: During GRES training, following \cite{gres, vpp_llava}, we merge the masks of all referred objects in a multi-target expression into a single combined mask. Accordingly, we compute a unified bounding box by taking the minimum and maximum coordinates of all associated object boxes, i.e., $(x_{\text{min}}, y_{\text{min}}, x_{\text{max}}, y_{\text{max}})$. For no-target samples, we set the ground truth mask to all zeros and define the bounding box as $(0, 0, 0, 0)$.}}

\noindent \textbf{Inference settings.}
In the inference phase, our models undergo two types of testing. Initially, we evaluate them in a standard setting where the training, validation, and test splits are derived from the same dataset (classic RES and GRES). We select the model for testing based on its highest gIoU score on the validation set. For instance, we train and test SSP-SAM on the RefCOCO dataset. Subsequently, to assess the open-vocabulary capability, following \cite{RISAM} we test the models on the test split of PhraseCut. These models are trained on RefCOCO, RefCOCO+, and RefCOCOg datasets, respectively. To prevent information leakage, no pre-training data is used in this experiment. More details can be found in \textbf{Section \ref{sec:open_vocab}}.

\subsection{Comparisons in Classic RES Scenario}
\label{exp_main}

\input{table/main_results}
\input{table/main_results_pr}

In Table~\ref{tab:res_sota}, we report the gIoU (\%) and cIoU (\%) of our proposed SSP-SAM and other state-of-the-art methods on four widely used RES benchmarks: RefCOCO, RefCOCO+, RefCOCOg, and ReferIt. 
  
The results clearly show that our proposed SSP-SAM framework outperforms both Non-SAM-based and SAM-based state-of-the-art methods across all splits of RefCOCO, RefCOCOg, and ReferIt (gIoU metric). Specifically, compared with the strongest Non-SAM-based method CGFormer on the RefCOCO dataset, SSP-SAM-336 achieves absolute gIoU gains of $2.44\%$, $3.25\%$, and $2.2\%$ on the val, testA, and testB splits, respectively.
In comparison with vanilla SAM and the SAM-based RISAM, our method yields substantial improvements, demonstrating the benefit of incorporating semantic-spatial prompts.
Notably, although Prompt-RIS also leverages prompts to guide SAM, its use of hand-crafted geometric cues limits its effectiveness. In contrast, our learned SSP provides more informative and adaptive guidance, maximizing the segmentation potential of SAM.

{In terms of the cIoU metric, our method consistently performs well on RefCOCO, RefCOCOg, and ReferIt. For RefCOCO+, where expressions are prohibited from using absolute location words (e.g., ``\textit{left}”, ``\textit{right}”), the effectiveness of our language attention adapter may be partially limited. Nevertheless, our method still achieves the best cIoU on the testA split of RefCOCO+. This result indirectly emphasizes the importance of spatial cues in referring expressions and, in turn, demonstrates the value of incorporating our language attention adapter to better capture such discriminative information when it is present.}

Although recent large models like LISA and GSVA, which incorporate SAM and stronger language backbones (e.g., Vicuna-7B), sometimes outperform our method on challenging subsets such as RefCOCO+ testB, our approach using a lightweight text encoder and less training data still achieves competitive or superior performance on datasets such as RefCOCO val and testA, demonstrating its effectiveness.

Table~\ref{tab:res_pr} supplements the gIoU and cIoU metrics by presenting Pr@0.5, Pr@0.7, and Pr@0.9 results on the RefCOCO val spilt for a more detailed analysis. These results, combined with observations in \textbf{Section~\ref{sec:gres}} and qualitative examples in \textbf{Section~\ref{sec:visualization}}, demonstrate that our model tends to produce high-quality masks rather than overly large or ambiguous ones. Even under the strict Pr@0.9 criterion, our method outperforms state-of-the-art approaches such as CGFormer and MagNet by up to 9\%. This explains why our cIoU scores may sometimes appear less impressive because cIoU favors larger masks, yet our model achieves superior visual and semantic alignment. In fact, neither gIoU nor cIoU fully capture the perceptual quality and precision of the predicted masks.

In general, the proposed SSP encoder effectively and seamlessly utilizes the strengths of CLIP and SAM to produce high-quality language-guided masks for the referent.

\subsection{Comparisons in GRES Scenario}
\label{sec:gres}
\input{table/gres}
\input{table/gres_pr}
In Table~\ref{tab:gres}, we report the GRES results on the gRefCOCO dataset.
Although our method is not specifically designed for the GRES setting, it naturally extends from classic RES scenarios—where each expression corresponds to a single target—to the more complex GRES task. Our approach achieves the best overall performance in terms of both gIoU and cIoU across all compared methods.
Given that GRES remains a relatively underexplored area, existing SAM-based models evaluated on this benchmark are limited and primarily rely on large MLLMs. Despite this, our method still demonstrates clear superiority over these approaches, highlighting its strong generalization and scalability. This further underscores the effectiveness and efficiency of our approach, even without extensive architectural modifications.

For the N-acc. metric, we follow the GRES protocol~\cite{gres}, which treats masks with fewer than 50 foreground pixels as all-negative. As shown in the table, our method achieves the highest N-acc. on gRefCOCO, highlighting its robustness and reliability in handling no-target scenarios.

As a supplement, we also present Pr@0.7, Pr@0.8, and Pr@0.9 on the val split (Table~\ref{tab:gres_pr}). Interestingly, although SSP-SAM reports slightly lower precision at Pr@0.7 compared to other methods, it significantly outperforms the others at higher thresholds. This suggests that our model tends to produce more precise and tightly aligned masks, which are especially beneficial in applications requiring fine-grained segmentation accuracy. Specifically, our Pr@0.9 surpasses CGFormer and ReLA by over 20\%. This improvements aligning well with the trend observed in classic RES benchmarks.

Overall, the proposed Semantic-Spatial Prompt effectively transfers CLIP's multi-modal understanding to SAM, fully exploiting its segmentation ability. Without any specialized design, it seamlessly generalizes to the GRES setting, highlighting the robustness and flexibility of our approach.

\subsection{Comparisons in Open-vocabulary Scenario}
\label{sec:open_vocab}
\input{table/phrase_cut}

Table~\ref{tab:zero-shot} reports the performance of the proposed method in zero-shot open-vocabulary setttings. The involved methods are trained on RefCOCO, RefCOCO+, and RefCOCOg (without pre-training strategy employed), and subsequently tested on the test set of the PhraseCut. We find that SSP-SAM achieves the best open-vocabulary performance across all three training sets. Specifically, SSP-SAM-336 achieves an gIoU of $28.29\%$ on PhraseCut test set when trained on RefCOCOg, which surpasses CRIS, LAVT and RISAM by absolute margins of $12.05\%$, $12.24\%$, and $5.82\%$ respectively. These results demonstrate that SSP-SAM fully leverages the pre-trained information from CLIP and SAM, exhibiting strong robustness in open-vocabulary scenarios compared to previous methods. Moreover, comparisons between RISAM and our proposed method suggest that leveraging Semantic-Spatial Prompts to fully exploit SAM's inherent segmentation capabilities could be more effective than distilling SAM's features into manually crafted student models.

\subsection{Time Cost Comparison}
\begin{table}[t]
\definecolor{lightblue}{RGB}{240, 248, 255} 
  \centering
  \renewcommand{\arraystretch}{1.2}
  \caption{{Comparison of the time cost with state-of-the-art methods on the testB split of RefCOCO dataset.}}
  \setlength\tabcolsep{10pt}
  {
    \begin{tabular}{lcccc}
    \toprule
    Method & Params. & Inference Time (ms) & Std. (ms) \\
    \midrule
    \textit{Layer-wise} DIT-B & 212.87M & 52.57 & 2.66 \\
    LISA-Vicuna-7B & 288.26M & 1324.46 & 98.98 \\
    \midrule
    \rowcolor{lightblue}SSP-SAM-224 & 18.37M & 26.68 & 2.10 \\
    \rowcolor{lightblue}SSP-SAM-336 & 18.73M & 53.00 & 1.26 \\
    \bottomrule
    \end{tabular}%
  }
  \label{tab:time_consumption}%
\end{table}%

{To better analyze the efficiency of SSP-SAM, we conduct experiments to measure the inference time and parameter complexity. All methods are evaluated on the RefCOCO testB split with batch size $1$ using a single NVIDIA A100 80GB GPU, which is necessary for MLLM-based methods like LISA-Vicuna-7B and ensures fairness. To avoid the impact of PyTorch cold start, the first $50$ samples are discarded. Then, $20$ random samples are repeatedly measured for $5$ runs, and we report the mean and standard deviation.}

{As shown in Table~\ref{tab:time_consumption}, SSP-SAM-224 requires only $18.37$M trainable parameters, which is much fewer than \textit{layer-wise} DIT-B and LISA-Vicuna-7B. In terms of inference speed, SSP-SAM-224 processes a sample in $26.68$ ms, nearly $2\times$ faster than \textit{layer-wise} DIT-B and over $50\times$ faster than LISA-Vicuna-7B. Even at higher resolution, SSP-SAM-336 maintains a comparable inference latency to \textit{layer-wise} DIT-B. For \textit{layer-wise} DIT, we report results using the officially released SAM-B backbone because the SAM-H version is not publicly available. Using SAM-H would likely further increase both parameter complexity and inference latency.}

{Taken together with the results in the \textbf{Classic RES Scenario}, \textbf{GRES Scenario}, and \textbf{Open-vocabulary Scenario}, these findings consistently demonstrate that SSP-SAM outperforms existing methods in all three key aspects: segmentation performance, lightweight parameterization, and inference efficiency, indicating the potential of SSP as an effective mechanism to guide SAM in real-world scenarios.}

\subsection{Backbone Generalization Beyond CLIP}
\begin{table}[t]
  \centering
  \caption{{Ablation on different SSP backbones for the SSP-SAM-224 framework on RefCOCO dataset. All models use SAM-H as the image encoder.}}
  \setlength\tabcolsep{18pt}
  \renewcommand{\arraystretch}{1.3}
  \begin{tabular}{lcccc}
    \toprule  
    \multirow{2}{*}{SSP Backbone} & \multicolumn{3}{c}{RefCOCO} \\
    & val & testA & testB \\
    \midrule
    CLIP     & \underline{76.24} & \underline{79.30} & 71.77 \\
    EVA-CLIP & \textbf{76.81}    & 79.23             & \underline{72.53} \\
    SigLIP2  & \textbf{76.81}    & \textbf{79.44}    & \textbf{72.86} \\
    \bottomrule
  \end{tabular}%
  \label{tab:ablation_backbone}
\end{table}

{Our SSP-SAM framework transfers multi-modal knowledge from a frozen vision-language model into semantic-spatial prompts to guide SAM. In our main experiments, the SSP encoder uses CLIP, which is also the most common backbone in RES~\cite{RISCLIP,DETRIS, RISAM, prompt_ris, remamber}. To further evaluate the generality of SSP encoder, we also conduct experiments with alternative backbones. Since these backbones are incompatible with our CLIP-based pretraining, all models are trained from scratch on RefCOCO using the ViT-B/16-224 configuration. Specifically, our experiments involve three SSP backbones: CLIP-ViT-B/16-224 (used in the main experiments), EVA-CLIP-ViT-B/16-224~\cite{eva-clip}, and SigLIP2 ViT-B/16-224~\cite{siglip2}.}

{As shown in Table~\ref{tab:ablation_backbone}, the performance is maintained or even improved with these backbones. For instance, on the RefCOCO testB split, replacing CLIP with SigLIP2 improves the gIoU from 71.77\% to 72.86\%, confirming strong adaptability. This demonstrates that SSP encoder can effectively transfer knowledge from various pre-trained vision-language models to guide SAM, highlighting the generality and scalability of our framework for future multi-modal integrations.}

\subsection{Ablation Study} 
\label{ablation}
\subsubsection{Training Strategy}
\label{training_strategy}
\begin{table}[t]
  \centering
  \caption{The ablation study of different training strategies on RefCOCO dataset. \textit{Aux.} is the abbreviation for the auxiliary task.
}
\setlength\tabcolsep{10pt}
  {
  \resizebox{\linewidth}{!}{
    \begin{tabular}{cccccc}
    \toprule
    \multirow{2}{*}{Pre-training} &\multirow{2}{*}{RES} &\multirow{2}{*}{Aux.} & \multicolumn{3}{@{}c@{}}{RefCOCO} \\
    & & &  val & testA & testB \\
    \midrule
     & \ding{51} & & 76.03 & 78.55 & 71.60  \\
     & \ding{51} & \ding{51} & 76.24 & 79.30 & 71.77 \\
    \ding{51} &\ding{51}&\ding{51}& \textbf{77.91} & \textbf{80.67} & \textbf{73.71}  \\
    \bottomrule
    \end{tabular}%
    }
    }
  \label{tab:Training_Strategy}%
  \end{table}

 The model's results of gIoU with different training strategies, based on the CLIP-ViT-B/16-224 backbone, are presented in Table~\ref{tab:Training_Strategy}. We can observe that when trained solely on the RES task, as indicated in Table~\ref{tab:res_sota} and Table~\ref{tab:Training_Strategy}, our method achieves comparable performance to RISAM. 
 To take advantage of the annotated bounding box data, we introduced the auxiliary task, which further enable the pre-training strategy. Upon introducing the auxiliary task, the performance of our proposed method has a notable improvement, surpassing both PolyFormer and VG-LAW on the RefCOCO val and testA splits. However, there is still room for further improvement. Inspired by existing methods such as RefTR, SeqTR, and PolyFormer, we adopt the pre-training strategy. As a result, our model achieves marked advancements, with an gIoU of $73.71\%$ on the RefCOCO's testB split. When considering Table~\ref{tab:res_sota}, our method outperforms peer models on the majority of datasets, with the exception of RefCOCO+, where a detailed analysis is provided in \textbf{Section~\ref{exp_main}}.

\subsubsection{Impact of freezing SAM Decoder}
\begin{table}[t]
\definecolor{lightblue}{RGB}{240, 248, 255} 
  \centering
  \caption{The ablation study on the impact of freezing SAM mask decoder on RefCOCO dataset. $*$ indicates the model with a frozen SAM decoder during training.}
\setlength\tabcolsep{10pt}
  {
  \resizebox{\linewidth}{!}{
    \begin{tabular}{lccccc}
    \toprule
    \multirow{2}{*}{Method} &\multirow{2}{*}{Freeze} & \multicolumn{3}{@{}c@{}}{RefCOCO} \\
     & & val & testA & testB \\
    \midrule
    SSP-SAM-224*& \ding{51} & 76.74 & 79.61 & 72.35  \\
    \rowcolor{lightblue}SSP-SAM-224& \ding{55}  & \textbf{77.91} & \textbf{80.67} & \textbf{73.71} \\
    \midrule
    SSP-SAM-336*& \ding{51} & 78.13 & 80.89 & 73.80  \\
    \rowcolor{lightblue}SSP-SAM-336& \ding{55} & \textbf{79.37} & \textbf{81.95} & \textbf{75.52} \\
    \bottomrule  
    \end{tabular}%
    }
    }
  \label{tab:Training_freeze_sam}%
  \end{table}

In Table~\ref{tab:Training_freeze_sam}, we report the effects of freezing the SAM decoder on segmentation results. The symbol $*$ in the table denotes the model configuration where the SAM decoder is frozen during training. We can observe that fine-tuning the SAM decoder after the first 20 epochs yields a 1\%-2\% performance increase on RefCOCO for both SSP-SAM-224 and SSP-SAM-336. For example, SSP-SAM-336 achieves $75.52\%$ on RefCOCO testB, which outperforms SSP-SAM-336* (frozen) by absolutely $1.72\%$. Combining with Table~\ref{tab:res_sota}, even with the SAM decoder fixed, our method (SSP-SAM-224* and SSP-SAM-336*) still achieves the best performance on val and testA of RefCOCO, and it attains performance on testB that is competitive with other state-of-the-art methods. Considering overall performance, we adopt the strategy that fine-tuning the SAM decoder after the first $20$ epochs as mentioned in \textbf{Section~\ref{implementation}}.

\subsubsection{Components}
\label{sec:ablation_components}
In Table \ref{tab:ablation}, we present the performance of the proposed framework with various modules on the RefCOCO dataset. To conserve computational resources, our experiments are conducted using the SSP-SAM-224 model. These components are prompt generator, visual attention adapter, and linguistic attention adapter, which are abbreviated as ``P.G.",``V.A.D." and ``L.A.D.", respectively. It is important to note that, in the absence of the P.G., the attention-adapted referent features $F_{SS}$ are directly projected by an MLP and forwarded to SAM as prompts. Several key observations can be summarized as follows.

Firstly, we observe that employing single modules in isolation, namely the L.A.D., V.A.D., and P.G., does not produce the optimal performance for our model. Specifically, it is important to clarify that using only P.G., without any enhancement to the attention mechanism, results in the direct utilization of the original CLIP features. As CLIP is trained at the image-text level and does not specialize in perceiving the semantics of individual regions, this approach leads to poor prompt quality. Consequently, it falls to fully leveraging SAM's inherent advantages in spatial awareness.

Secondly, when comparing the model's performance of only V.A.D. and only L.A.D., we find that optimizing linguistic attention over visual attention results in the generation of higher-quality prompts. This finding also supports the argument presented in \textbf{Section~\ref{exp_main}} that lack of the discriminative linguistic features may result in sub-optimal performance. We also note that focusing exclusively on adjusting visual attention can substantially refine the potential locations of the referent within the prompts. Consequently, even in scenarios where there is an absence of discriminative linguistic information, such as positional details, there is a significant enhancement in performance compared to utilizing the original CLIP features.

Thirdly, the combination of L.A.D. and V.A.D. leads to a significant performance boost, which is especially evident in scenarios featuring a greater diversity of object categories, such as on the RefCOCO testB. Nevertheless, using the adjusted features directly as prompts for SAM does not fully utilize its segmentation capabilities. This may stem from to SAM's original design, which is tailored to respond to geographical prompts. Thus, our generated prompts require further refinement through the integration of learnable special tokens designed to effectively extract Semantic-Spatial information related to the referent. Such prompts are more effectively suitable for the mask decoder of SAM. Upon incorporating all the modules, we can observe a significant enhancement in performance. This demonstrates that our prompt encoder effectively integrates CLIP's multi-modal pre-trained knowledge into SAM through the formulation of Semantic-Spatial prompts. As a result, SAM has seamlessly acquired the capability to handle RES with promising performance.
\begin{table}[t]
  \centering
  \caption{The ablation study of different modules on RefCOCO dataset. ``P.G.", ``V.A.D." and ``L.A.D." are the abbreviation for Prompt Generater, Visual Attention Adapter and Linguistic Attention Adapter, respectively.
}
\setlength\tabcolsep{10pt}
  {
  \resizebox{\linewidth}{!}{
    \begin{tabular}{cccccc}
    \toprule
    \multirow{2}{*}{P.G.}&\multirow{2}{*}{V.A.D.} &\multirow{2}{*}{L.A.D.} & \multicolumn{3}{@{}c@{}}{RefCOCO} \\
    & & &  val & testA & testB \\
    \midrule
    \ding{51}& & & 42.30 & 43.13 & 41.18  \\
    &\ding{51}& & 61.27 & 63.83 & 58.97 \\
    &&\ding{51}& 66.09 & 70.24 & 61.78  \\
    
    & \ding{51} &\ding{51} & 67.87 & 71.24 & 63.11 \\
    \ding{51}&\ding{51}&\ding{51}& \textbf{77.91} & \textbf{80.67} & \textbf{73.71}  \\
    \bottomrule
    \end{tabular}%
    }
    }
  \label{tab:ablation}%
  \end{table}

\subsubsection{Prompt Generator Type}
\label{sec:ablation_pg_type}
{As discussed in the previous Components ablation (\textbf{Section~\ref{sec:ablation_components}}), using a simple MLP to generate prompts from $F_{SS}$ leads to suboptimal performance. Building on this, we further compare three types of prompt generators on RefCOCO: MLP, DETR-like decoder, and our encoder-based default prompt generator. As shown in Table~\ref{tab:pg_type_ablation}, the encoder-based default prompt generator concatenates the learnable tokens with $F_{SS}$ and feeds them into a transformer encoder, allowing $F_{SS}$ to further enhance themselves while interacting with the learnable tokens. This produces more informative prompts and better performance compared to an MLP or a DETR-like decoder, where such self-enhancement of $F_{SS}$ is not possible.
}

\definecolor{lightblue}{RGB}{240, 248, 255} 

\begin{table}[t]
  \centering
  \caption{{Ablation study of different types of prompt generator on RefCOCO dataset. ``P.G.'' stands for Prompt Generator.}}
  \setlength{\tabcolsep}{11pt}
  \renewcommand{\arraystretch}{1.2} 
  \resizebox{\linewidth}{!}{%
  \begin{tabular}{lcccc}
    \toprule
    \multirow{2}{*}{P.G. Type} & \multicolumn{3}{c}{RefCOCO} \\
    & val & testA & testB \\
    \midrule
    MLP & 67.87 & 71.24 & 63.11 \\
    DETR-like decoder & 72.54 & 76.13 & 67.82 \\
    \rowcolor{lightblue}Default (encoder-based) & \textbf{77.91} & \textbf{80.67} & \textbf{73.71} \\
    \bottomrule
  \end{tabular}%
  }
  \label{tab:pg_type_ablation}
\end{table}

\subsubsection{Effect of Gaussian-like Mapping}
\label{sec:ablation_gaussian}

{Following VLTVG~\cite{vltvg} and TransCP~\cite{transcp}, the Gaussian-like mapping in Eq.~\ref{word-level-select} further enhances the discrimination of salient objects from background by sharply increasing the weights of text-aligned regions while suppressing unrelated features. To evaluate its necessity, we conduct an ablation study on SSP-SAM-224. Since this mapping is also applied during pretraining, we train the model from scratch to avoid any influence from pretrained weights. As shown in Table~\ref{tab:gaussian}, removing the Gaussian-like mapping (SSP-SAM-224$^\dagger$) leads to a performance drop across all RefCOCO splits (val: 75.38\%, testA: 78.54\%, testB: 70.79\%), while including it consistently improves results (val: 76.24\%, testA: 79.30\%, testB: 71.77\%), validating its role in increasing sensitivity to salient objects and suppressing irrelevant context.}

\definecolor{lightblue}{RGB}{240, 248, 255} 
\begin{table}[t]
  \centering
  \caption{{Ablation study of Gaussian-like mapping on RefCOCO dataset. ``G.L.M.'' indicates the presence of the Gaussian-like mapping and ``$\dagger$'' indicates its removal.}}
  \setlength{\tabcolsep}{11pt}
  \renewcommand{\arraystretch}{1.2} 
  \resizebox{\linewidth}{!}{%
  \begin{tabular}{lccccc}
    \toprule
    \multirow{2}{*}{Method} & \multirow{2}{*}{G.L.M.} & \multicolumn{3}{c}{RefCOCO} \\
    & & val & testA & testB \\
    \midrule
    SSP-SAM-224$^{\dagger}$ & & 75.38 & 78.54 & 70.79 \\
    \rowcolor{lightblue}SSP-SAM-224 & \ding{51} & \textbf{76.24} & \textbf{79.30} & \textbf{71.77} \\
    \bottomrule
  \end{tabular}%
  }
  \label{tab:gaussian}
\end{table}


\subsubsection{Number of Prompt Generator Layers}
\label{sec:num_layer}

\begin{table}[t]
  \centering
  \caption{The ablation study for the numbers of transformer encoder layers in the prompt generator. We train the model on RefCOCO dataset and test on RefCOCO and the open-vocabulary PhreaseCut.}
\setlength\tabcolsep{12pt}
  {
  \resizebox{\linewidth}{!}{
    \begin{tabular}{ccccc}
    \toprule
    \multirow{2}{*}{Layers}&  \multicolumn{3}{c}{RefCOCO} & \multicolumn{1}{c}{PhraseCut}\\
      &  val & testA & testB & test\\
    \midrule
    1& 74.57 & 77.04 & 69.45 & 23.48  \\
    2& 75.56 & 78.16 & 71.64 & \textbf{25.05}  \\
    4& 76.27 & 78.75 & 71.61 & 24.81  \\
    6& 76.24 & \textbf{79.30} & 71.77 & 24.66  \\
    8& \textbf{76.43} & 79.18 & \textbf{72.14} & 24.59  \\
    \bottomrule
    \end{tabular}%
    }
    }
  \label{tab:supple_num_layers}%
  \end{table}
  
Table~\ref{tab:supple_num_layers} illustrates the impact of varying the number of transformer layers in the prompt generator. Since the pre-trained and fine-tuned models must share the same architecture (e.g., number of transformer layers), we report results using SSP-SAM-224 trained and evaluated on RefCOCO without pre-training for clarity. 
To better understand SSP-SAM, we additionally report the performance on the PhraseCut test dataset within the open-vocabulary setting. 
The results appear mixed, with no clear trend as the number of layers increases. Overall, our method exhibits stable performance on both RefCOCO and PhraseCut, suggesting robustness to this architectural choice. Given a slight improvement on RefCOCO and a marginal drop on PhraseCut, we select $6$ transformer layers as a balanced setting across datasets, also considering computational efficiency.

\subsubsection{Number of [RES] tokens}
\label{sec:num_ris}
\begin{table}[t]
  \centering
  \caption{The ablation study for the numbers of [RES] tokens in the prompt generator. We train the model on RefCOCO dataset and test on RefCOCO and the open-vocabulary PhreaseCut.}
\setlength\tabcolsep{12pt}
  {
  \resizebox{\linewidth}{!}{
    \begin{tabular}{ccccc}
    \toprule
    \multirow{2}{*}{Number}&  \multicolumn{3}{@{}c@{}}{RefCOCO} & \multicolumn{1}{@{}c@{}}{PhraseCut}\\
      &  val & testA & testB & test\\
    \midrule
    2& 76.21 & 78.68 & 71.88 & 25.22  \\
    16& \textbf{76.56} & 78.81 & 71.82 & \textbf{25.31}  \\
    32& 76.24 & 78.88 & 72.07 & 25.39 \\
    128& 76.24 & \textbf{79.30} & 71.77 & 24.66  \\
    256& 76.47 & 79.01  & \textbf{72.41} & 24.17  \\
    \bottomrule
    \end{tabular}%
    }
    }
  \label{tab:supple_num_res_tokens}%
  \end{table}

The ablation study on the number of RES tokens is presented in Table~\ref{tab:supple_num_res_tokens}. We follow the same experimental settings as those detailed in \textbf{Section~\ref{sec:num_layer}} and observe a similar phenomenon to that described therein. As the number of tokens increases, our method demonstrates relatively stable gIoU on both RefCOCO and PhraseCut, which also indicates the robustness of our approach to this parameter. Considering a slight decrease on PhraseCut and a fluctuating rise on RefCOCO with the increase in token numbers, we empirically choose $128$ as the final number of special tokens to balance performance across both scenarios.

\section{Discussion}
\label{sec:discussion}

{RES relies on both semantic and spatial cues, and SSP-SAM is designed to efficiently leverage spatial information when it is present. On RefCOCO+, a dataset with few explicit spatial expressions, the relative advantage of SSP-SAM is smaller. To examine how spatial information contributes to segmentation performance, we conduct a small-scale experiment on 25 randomly selected RefCOCO+ testB expressions. In this experiment, we preserve the original referring expressions as much as possible and add minimal positional information (e.g., “yellow dog” → “yellow dog on the far right”) to create augmented expressions. The model used in this experiment is trained on RefCOCO+, where explicit spatial terms rarely appear in the training data.}

As shown in Table~\ref{tab:ssp_spatial}, SSP-SAM-336 improves from 76.53\% to 80.88\% gIoU and from 68.79\% to 72.70\% cIoU after incorporating these additional spatial cues. These results demonstrate that, even though the model has limited exposure to explicit spatial terms during training, it can efficiently exploit spatial information when available. This observation underscores the value of integrating semantic and spatial knowledge in RES and highlights the flexibility and effectiveness of SSP in guiding SAM for RES.
\definecolor{lightblue}{RGB}{240, 248, 255} 
\begin{table}[h]
    \centering
    \caption{{Comparison of SSP-SAM under different spatial cues settings on 25 random samples from RefCOCO testB split.}}
    \setlength\tabcolsep{11pt} 
    \renewcommand{\arraystretch}{1.3} 
    \small
    \begin{tabular}{lcccc}
        \toprule
        \multirow{2}{*}{Method} & \multicolumn{2}{c}{w/o spatial cues} & \multicolumn{2}{c}{w/ spatial cues} \\
        & gIoU & cIoU & gIoU & cIoU \\
        \midrule
        SSP-SAM-224 & 70.63 & 60.49 & 73.23 & 62.99 \\
        SSP-SAM-336 & 76.53 & 68.79 & 80.88 & 72.70 \\
        \bottomrule
    \end{tabular}
    \label{tab:ssp_spatial}
\end{table}


\section{Qualitative Results}
\label{sec:visualization}
\begin{figure*}[!t]
    \centering
    \includegraphics[width=1\linewidth]{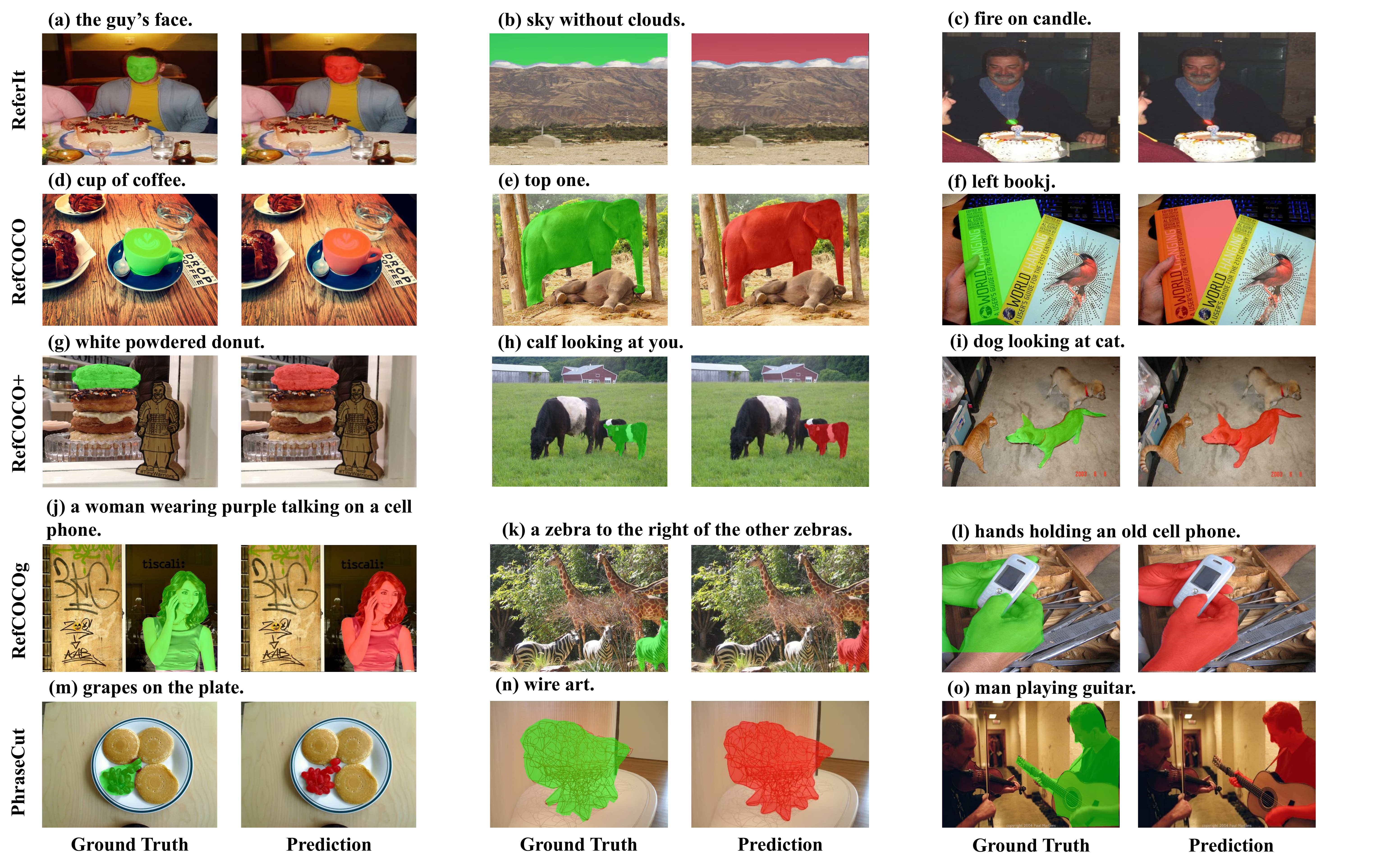}
    \caption{Qualitative segmentation results of SSP-SAM on classic RES datasets (RefCOCO, RefCOCO+, RefCOCOg, ReferIt) and the open-vocabulary PhraseCut dataset. SSP-SAM generates masks with sharper boundaries and finer details. Predicted masks are shown in red, and ground truth in green.}
    \label{fig:visualization_results}
\end{figure*}

\begin{figure*}[!h]
    \centering
    \includegraphics[width=1\linewidth]{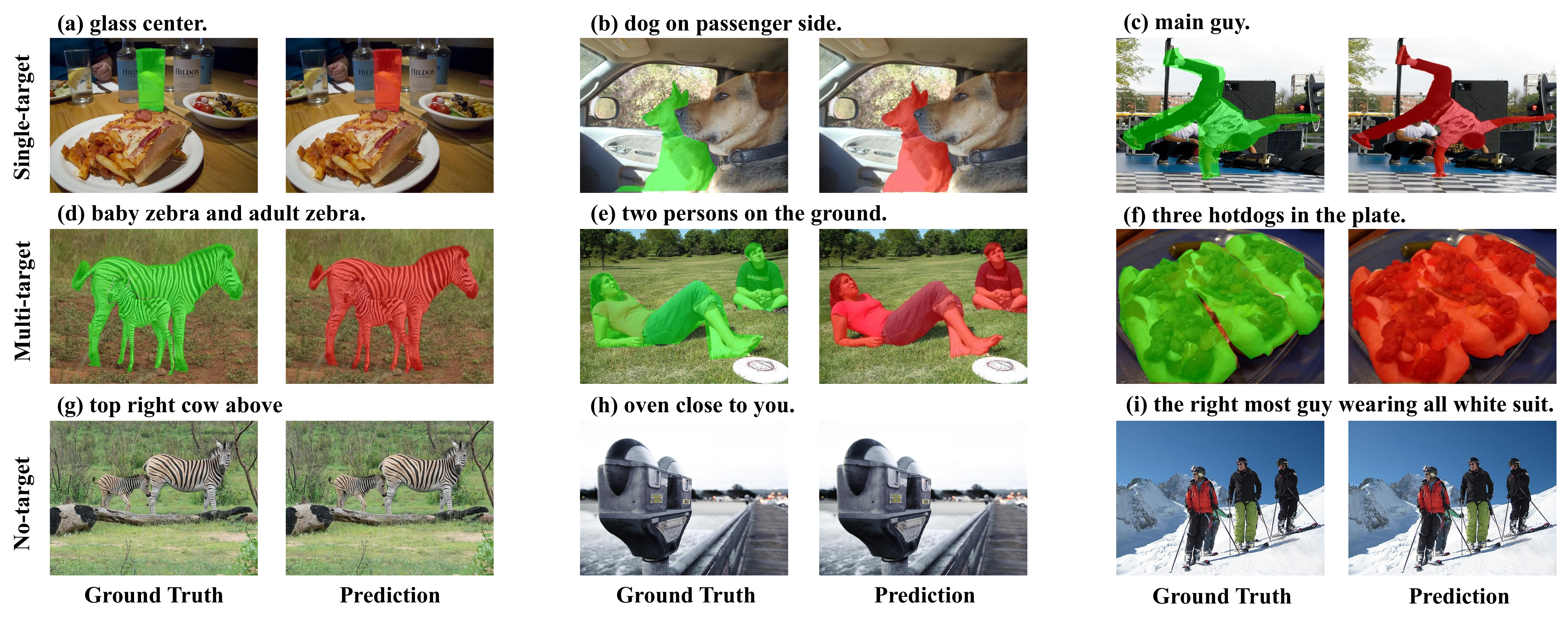}
    \caption{Qualitative results of SSP-SAM on the gRefCOCO dataset under the GRES scenario. Our method produces segmentation masks with sharper edges and finer details. Predicted masks are shown in red, and ground truth masks are shown in green.}
    \label{fig:visualization_gres}
    \vspace{-2.5mm}
\end{figure*}

\begin{figure*}[!t]
    \centering
    \includegraphics[width=1\linewidth]{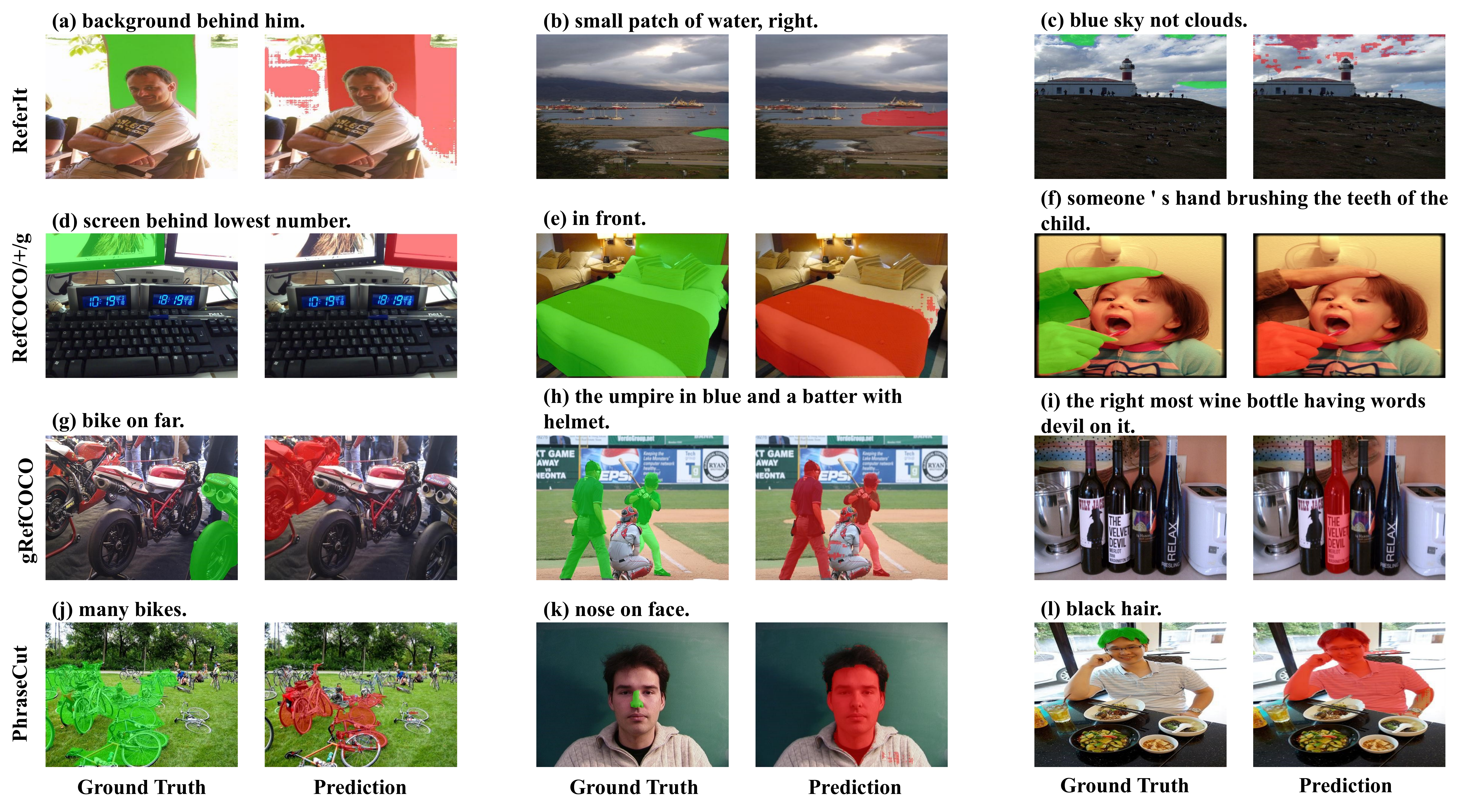}
    \caption{Failure cases of SSP-SAM on classic RES datasets (RefCOCO, RefCOCO+, RefCOCOg, ReferIt), the Generalized RES dataset (gRefCOCO), and the open-vocabulary PhraseCut dataset. Predicted masks are shown in red, and ground truth in green.}
    \label{fig:visualization_failure}
\end{figure*}

\begin{figure}[!h]
    \centering
    \includegraphics[width=1\linewidth]{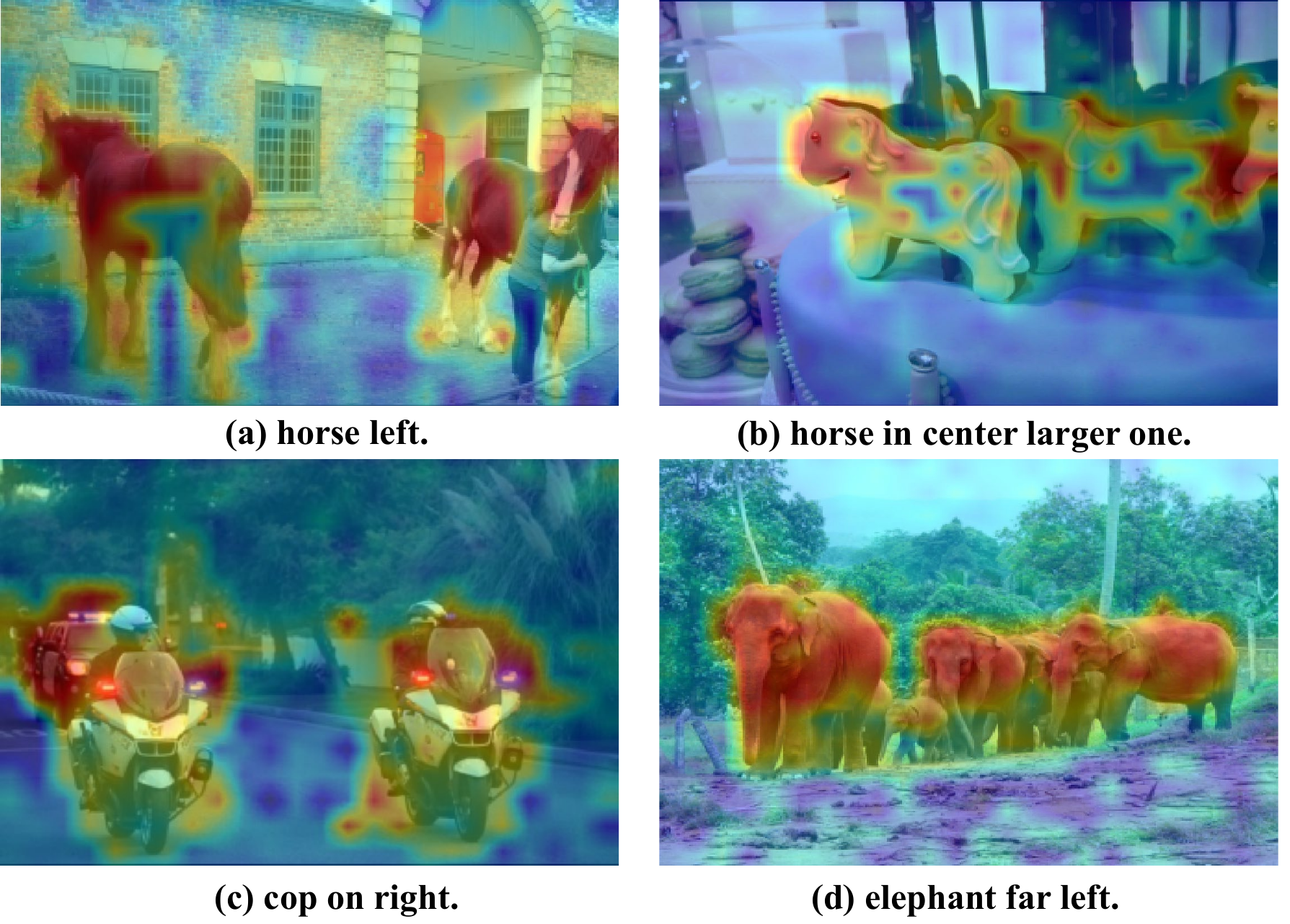}
    \caption{Qualitative results of the adapted visual attention. The salient objects within the images receive higher attention scores through the visual attention adapter.}
    \label{fig:visualization_attn_visual}
    \vspace{-2.5mm}
\end{figure}
\begin{figure}[!h]
    \centering
    \includegraphics[width=1\linewidth]{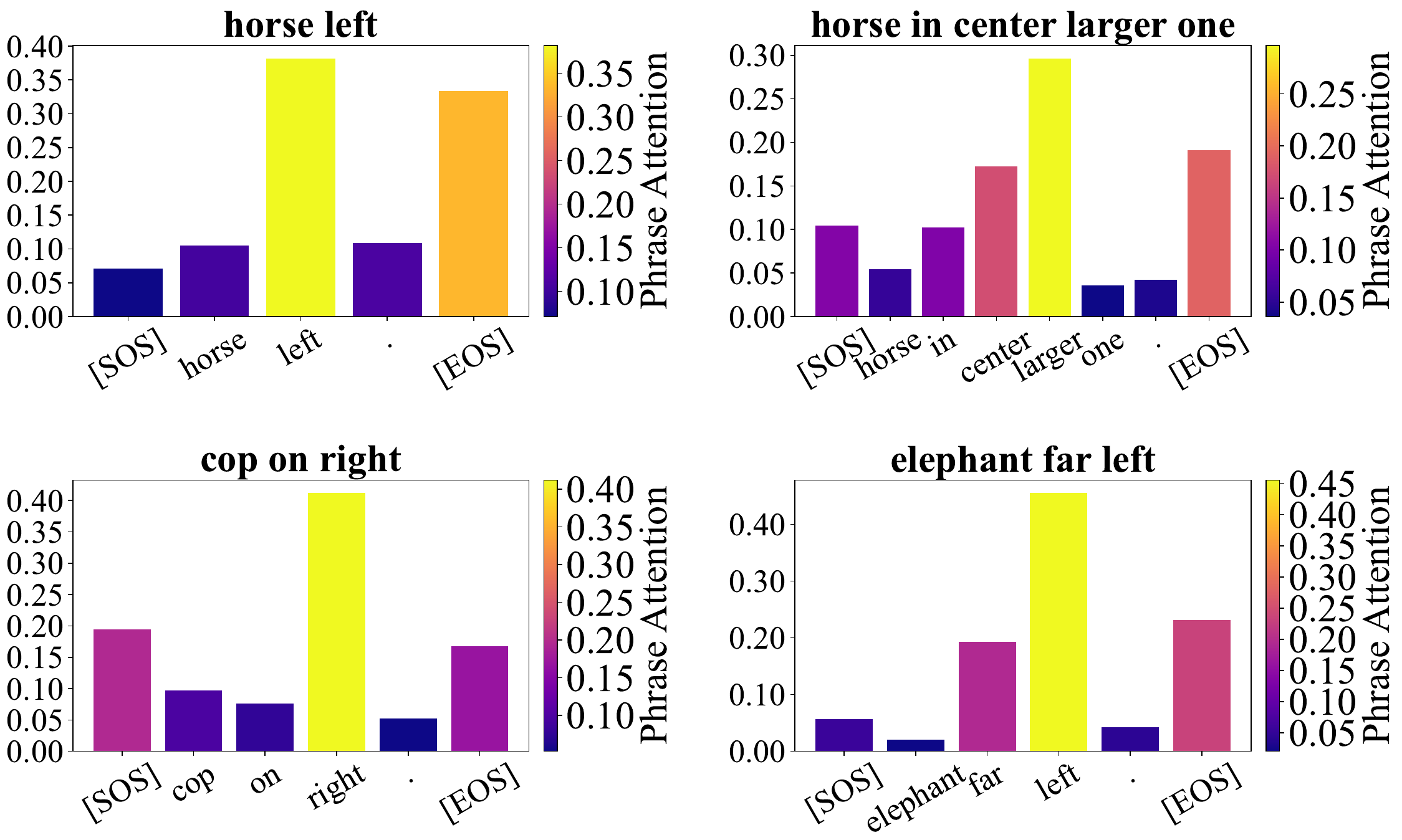}
    \caption{Qualitative results of the adapted linguistic attention. The discriminative phrases within the referring expressions receive higher attention scores through the linguistic attention adapter.}
    \label{fig:visualization_attn_linguistic}
    \vspace{-2.5mm}
\end{figure}

\subsection{Qualitative Results on Classic RES}
In Fig.~\ref{fig:visualization_results}, we present our qualitative segmentation results alongside the ground truth masks for comparison. From Fig.~\ref{fig:visualization_results}, it is evident that our model successfully segments the objects referred in the given expressions. Surprisingly, our segmentation results outperform the ground truth in some cases, offering masks with sharper edges and finer details. For instance, in case (a), our method accurately segments the entire face, including fine details, while the ground truth mask appeared to be rough, omitting areas such as the forehead and ears. A similar phenomenon is observed in other cases such as the clouds of the sky (case (b)), the handle of a coffee cup (case (d)), the hand on the book (case (f)), and the tail of an elephant (case (e)). In these cases, our predicted masks accurately segment the referent while excluding the background, whereas the ground truth masks include some background elements. Furthermore, our method provides accurate segmentation masks for small objects (case (c) and case(m)), which are traditionally a challenging aspect of segmentation tasks. 

Lastly, we selected the model trained on RefCOCOg, without pre-training the SSP encoder, for evaluation of its performance on the open-vocabulary PhraseCut test set, as depicted in case (m), (n) and (o). The phenomenon previously mentioned is also observed in the open-vocabulary settings, where SSP-SAM maintains its robust performance despite encountering objects unseen during training.

In summary, all these visualizations demonstrate that the proposed Semantic-Spatial Prompt encoder effectively utilizes CLIP’s vision-language correspondence and SAM's inherent superior segmentation capabilities to delivers high-quality segmentation results based on language conditions.

\subsection{Qualitative Results on GRES}
We present qualitative segmentation results under the GRES scenario in Fig.~\ref{fig:visualization_gres}, alongside ground-truth masks for comparison. Three representative examples are shown for each of the single-target, multi-target, and no-target (empty) cases from the gRefCOCO dataset.

For the single-target setting, our model demonstrates consistent performance with results observed in classic RES tasks. Notably, in case (c), the predicted mask is significantly more precise than the corresponding ground truth.

For the multi-target setting, SSP-SAM effectively captures the semantics of referring expressions and transfers both semantic and spatial cues from CLIP to SAM via our proposed SSP. This enables the model to leverage SAM’s strong segmentation and generalization capabilities. As shown in cases (e) and (f), whether the expression refers to two or more targets, our method produces high-quality masks with finer boundaries than the ground truth. For the no-target case, our model produces strictly all-negative masks, with no foreground pixels predicted. This result is consistent with the quantitative results in \textbf{Section~\ref{sec:gres}}.

Overall, without task-specific architectural modifications, our method naturally extends to GRES through SSP, demonstrating superior robustness, effectiveness, and flexibility.

\subsection{Failure Cases Study}
\label{subsec:failure}

To provide a more balanced analysis, we further investigate failure cases of SSP-SAM, as illustrated in Fig.~\ref{fig:visualization_failure}. Several patterns can be observed.

Firstly, when expressions lack discriminative language, the SSP encoder may produce unreliable prompts, resulting in incorrect masks (cases (a), (b), (g), and (e)).

Secondly, with complex expressions involving negations or subtle structures, the model may misinfer target positions. This is more pronounced when multiple objects are present (case (h)) or when the described object exists but some details slightly mismatch the referring language (case (i)).

Lastly, in the open-vocabulary setting, additional challenges arise from unseen patterns. Common failures include large number multiple-target cases (case (j)) and partial object descriptions (cases (k) and (l)), where masks are often inaccurate.

\subsection{Visualization of Adapted Attention}
To better understand the proposed method, we show the attention scores after the proposed attention adapter in Fig.~\ref{fig:visualization_attn_visual} and Fig.~\ref{fig:visualization_attn_linguistic}. We can observe that in visual attention adapter, the salient objects, for example, the horses in case (a), or the elephants in case (d) are highlighted by assigning the higher attention scores than the context. The adapted visual attention effectively narrows down the potential regions of the referent within the attention map, allowing for an initial integration of spatial information with the semantics of the referent. 
{Meanwhile, the linguistic attention adapter highlights discriminative relations or attributes in the expressions. Instead of enforcing a single dominant token, it can assign high weights to multiple words or phrases, such as \textit{left}, \textit{center}, \textit{larger}, or \textit{right}, thereby enhancing the integration of spatial and semantic information.}
These attention-refined features result in the high quantity of Semantic-Spatial Prompts, which effectively guide SAM in generating language-conditioned segmentation masks. As also demonstrated in Fig.~\ref{fig:visualization_results} and Fig.~\ref{fig:visualization_gres}, the proposed method successfully generates impressive segmentation masks for the referent through the Semantic-Spatial Prompts.

\section{Conclusion}
In this paper, we propose a straightforward one-stage framework SSP-SAM with the Semantic-Spatial Prompt (SSP) encoder. The SSP encoder empowers SAM to seamlessly leverage the vision-language correspondence from CLIP for tackling RES tasks, including classic RES and GRES. It effectively transforms the given image and language into semantically enriched and spatially detailed prompts using its visual and linguistic attention adapter. The visual attention adapter highlights salient objects, achieving preliminary integration of spatial information relevant to the referent's semantics.
The referent features are further refined by the adapted linguistic features, which dynamically focus on discriminative phrases via the linguistic attention adapter. The SSP, produced by a prompt generator, effectively captures cues for the referent within these attention-adapted features, guiding SAM to segment the target referred to by natural language.
Extensive experiments conducted on various RES and GRES benchmarks, as well as under open-vocabulary settings, demonstrate the superiority of SSP-SAM in effectively leveraging the inherent potential of SAM. Furthermore, the high quality of the predicted masks highlights the reliability and robustness of our approach.

In future work, we plan to: (1) enhance performance in open-vocabulary scenarios; (2) incorporate more training data and adopt stronger visual and textual encoder backbones; and (3) improve overall performance across varying precision levels, while maintaining the ability to produce high-quality segmentation masks.

\section*{Acknowledgments}
This work was supported by National Natural Science Foundation of China (Grant No. 62425603) and Basic Research Program of Jiangsu Province (Grant No. BK20243018).

\bibliographystyle{IEEEtran}
\bibliography{refer}

\end{document}

%% file: section/intro.tex
\begin{figure}[t]
    \centering
    \includegraphics[width=1\linewidth]{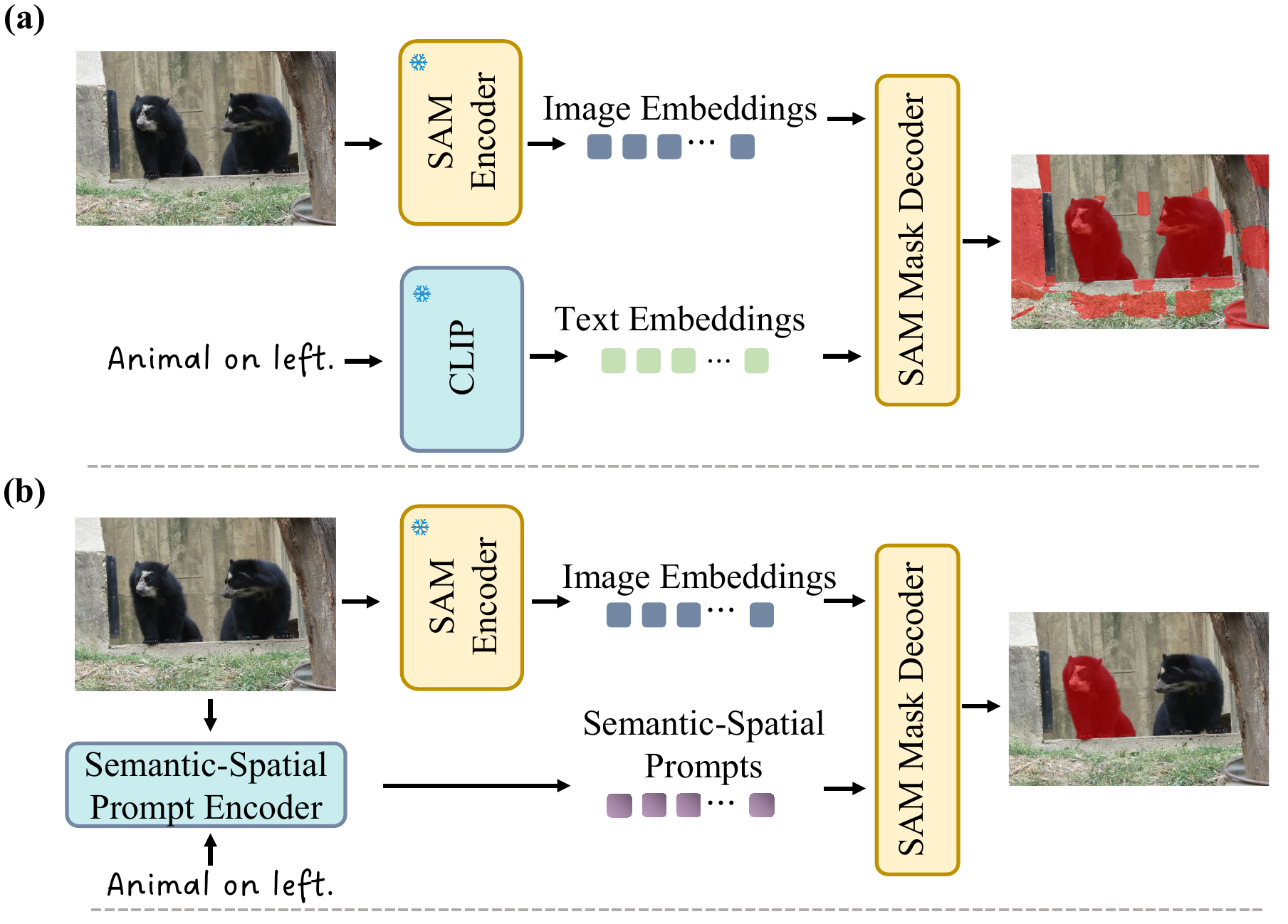}
    \caption{(a) SAM struggles with text prompts, especially the free-form languages in the Referring Expression Segmentation, resulting in poor segmentation mask. (b) SSP-SAM seamlessly transform images and languages to Semantic-Spatial Prompts to guide SAM for referent segmentation.}
    \label{fig:illustration}
\end{figure}

\begin{figure}[t]
    \centering
    \includegraphics[width=1\linewidth]{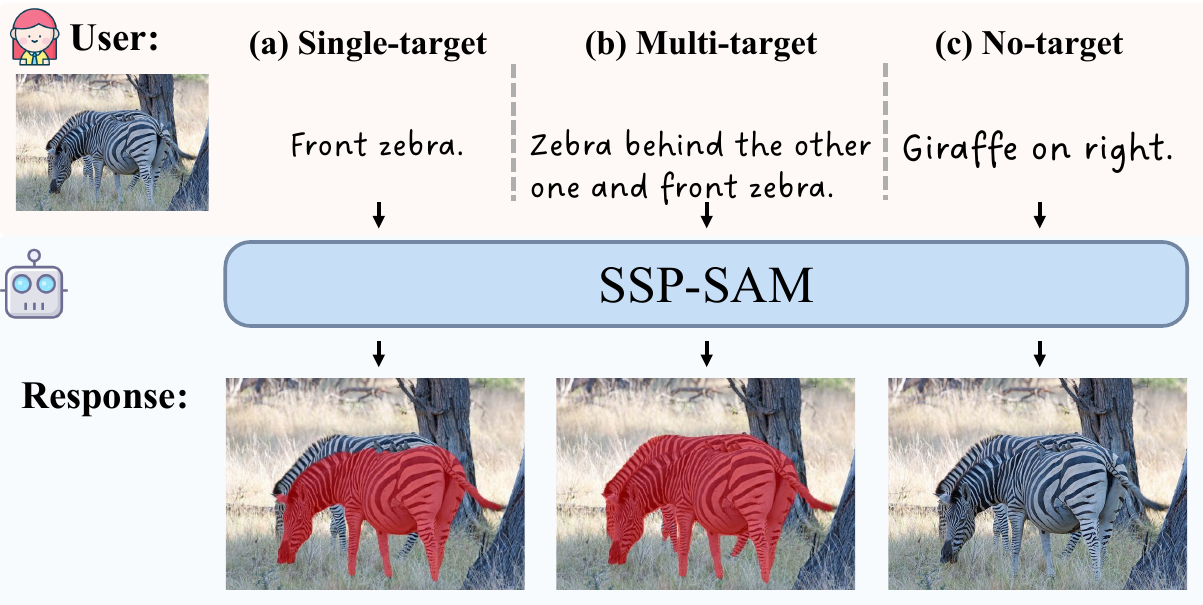}
    \caption{Illustration of referring expression segmentation (RES) settings. (a) Classic RES focuses on single-target segmentation. (b) and (c) Generalized RES (GRES) introduces multi-target and no-target scenarios, which require not only a comprehensive understanding of all objects in the scene but also a contextual interpretation of both the image and the referring expression.}
    \label{fig:ssp_sam_task}
    \vspace{-5mm}
\end{figure}

\IEEEPARstart{R}{eferring} Expression Segmentation (RES) aims to segment objects referred by free-form natural language expressions (known as the referent) \cite{tcsvt_crscnet, tcsvt_refersam, tcsvt_bms, tcsvt_pks, cris, lavt, cgformer, polyformer, vg-law}. It has numerous real applications, such as robot-human interaction \cite{seg_robot_1} and image editing \cite{IMGEDIT_RIS}. Compared to conventional semantic segmentation based on predefined categories \cite{ctnet, visual_incontext}, RES is more challenging, as it requires understanding fine-grained multimodal cues such as attributes and object relationships \cite{polyformer, vg-law}.

Recently, the Segment Anything Model (SAM) \cite{sam} has demonstrated impressive segmentation performance using over one billion image–mask pairs and supporting geometric prompts like points, boxes, and masks. However, SAM shows limited capability in understanding natural language, and thus struggles with segmenting targets described by free-form expressions \cite{sam_clip, vrp_sam, PPT, open-vocabulary-sam, semiSAM, DIT-H}. As shown in Fig.~\ref{fig:illustration}(a), the expression “\textit{Animal on left.}” leads SAM to include irrelevant objects, exposing its poor language ability.

To address this, recent efforts have incorporated linguistic guidance into SAM, primarily through pre-SAM and post-SAM paradigms \cite{SAM_Survey}. Both are essentially two-stage pipelines: pre-SAM \cite{fgvp, li2023clip_surgery} leverages external grounding models to generate geometric prompts, while post-SAM \cite{tas, semiSAM} produces mask proposals that are subsequently re-ranked using cross-modal models like CLIP \cite{clip}. Although these approaches achieve some gains, they decouple language understanding from mask prediction, limiting fine-grained alignment. More recently, alternative paradigms have emerged, including one-stage methods that perform end-to-end integration of SAM features \cite{RISAM, DIT-H, tcsvt_refersam, PPT} {(e.g., Prompt-RIS \cite{prompt_ris}, which relies on CLIP-generated geometric prompts, and EVF-SAM \cite{evf-sam}, which employs an early-fusion encoder but requires heavy training)}, as well as approaches that combine Multimodal Large Language Models (MLLMs) with SAM to strengthen language understanding and reasoning in RES \cite{LISA, gsva, refsam, LaSagnA}. 
{While these newer approaches mitigate some limitations of two-stage pipelines, they often still rely on geometric prompts, incur greater architectural complexity, and require large-scale training. Table~\ref{tab:sam_intro_summary_type} summarizes these representative methods.
}

\begin{table*}[t]
\centering
\scriptsize
\renewcommand{\arraystretch}{1}
\caption{{Summary of Representative SAM-based RES methods.}}
\begin{tabular}{@{}p{0.15\textwidth} p{0.15\textwidth} p{0.65\textwidth}@{}}
\toprule
Method & Type & Core Contribution \\
\midrule
Grounded SAM \cite{grounded_sam} & Pre-SAM & Generates language-driven prompts from external models to guide SAM. \\
CLIP Surgery \cite{li2023clip_surgery} & Pre-SAM & Samples points from attention regions to refine SAM mask proposals. \\
TAS \cite{tas} & Post-SAM & Generates mask proposals via SAM and selects the best using CLIP cross-modal matching. \\
Semi-SAM \cite{semiSAM} & Post-SAM & Refines SAM-generated masks using CLIP to improve alignment with text. \\
DIT \cite{DIT-H} & One-stage & Fine-grained layer-wise adjustments to SAM for improved segmentation. \\
Prompt-RIS \cite{prompt_ris} & One-stage & Text-guided SAM, yet still heavily dependent on CLIP-generated geometric prompts. \\
EVF-SAM \cite{evf-sam} & One-stage & Leverage early-fusion encoder for SAM with multi-modal prompts, yet heavily training-dependent. \\
LISA \cite{LISA} & MLLM-based & Integrates SAM features into MLLMs to decode language-conditioned masks. \\
\bottomrule
\end{tabular}
\label{tab:sam_intro_summary_type}
\end{table*}

These challenges become more pronounced in the Generalized RES (GRES) \cite{gres} setting, which extends classic RES by allowing expressions to refer to single, multiple, or no targets. As shown in Fig.~\ref{fig:ssp_sam_task}, expressions like “\textit{Zebra behind the other one and front zebra}” involve multiple targets, while “\textit{Giraffe on right}” yields a no-target case. GRES introduces greater ambiguity and variability, highlighting the need for unified models that integrate language and vision more deeply and generalize across diverse grounding scenarios.

Toward this end, we propose SSP-SAM, a one-stage, lightweight approach.
Our method leverages a Semantic-Spatial Prompt (SSP) encoder to capture rich semantic and spatial cues from both language and vision, enabling SAM to produce high-quality, language-conditioned masks for both classic RES and GRES scenarios without requiring specialized modifications. As illustrated in Fig.\ref{fig:illustration}(b) and Fig.\ref{fig:ssp_sam_task}, our method segments fine-grained referents (e.g., \textit{“Animal on left.”} or \textit{“Zebra behind the other one and front zebra.”}) with clear boundaries while excluding irrelevant content.

Specifically, we use CLIP to extract image and text features in the SSP encoder, as it provides strong performance in multi-modal learning. The core design of our prompt encoder is an attention adapter, illustrated in Fig.\ref{fig:framework} and Fig.\ref{fig:attn_adapter}, which consists of visual and linguistic components. The visual attention adapter first selects visual features at the sentence level and then generates improved attention scores at the word level to enhance features of salient objects, thus integrating spatial information into the semantic representation of the referent. The linguistic attention adapter dynamically adjusts the importance of language tokens to emphasize discriminative phrases, such as relationships and attributes. After adaptation, the refined language features act as a soft mask with semantic cues that select relevant visual regions via an element-wise product. This mechanism further enhances the integration of semantic and spatial information. Finally, to produce SSPs compatible with SAM's mask decoder, we introduce learnable prompt tokens that extract referent cues through a prompt generator. 
{Since high-quality mask annotations are costly and limited, we incorporate the Referring Expression Comprehension (REC) task as an auxiliary task during training to better exploit available bounding box annotations. This auxiliary supervision helps improve the quality of the generated masks by encouraging the model to leverage both mask and bounding box cues effectively.}

With this design, SAM is effectively empowered to process free-form natural language by leveraging SSPs derived from CLIP features. Our framework achieves state-of-the-art performance on both RES and GRES benchmarks, and generalizes well to challenging open-vocabulary scenarios on PhraseCut dataset \cite{phrasecut}. The superior quality of our generated masks is especially evident at high overlap thresholds, such as Pr@0.9, reflecting precise and reliable segmentation.

In summary, our main contributions are as follows:
\begin{itemize}
\item We design a Semantic-Spatial Prompt encoder that fuses semantic and spatial information from language and vision to guide SAM in generating accurate, language-conditioned masks.

\item {We develop SSP-SAM, a simple lightweight one-stage framework combining SAM and CLIP, effective for both classic RES and generalized RES without extra modifications.}

\item Our method achieves state-of-the-art results on RES and GRES benchmarks and shows improved zero-shot open-vocabulary performance, producing high-quality masks measured by strict metrics like Pr@0.9.
\end{itemize}

%% file: section/related_work.tex
\subsection{Referring Expression Segmentation}
Referring Expression Segmentation (RES) aims to generate pixel-level masks for objects described by natural language, and is closely related to Referring Expression Comprehension (REC) \cite{liu2022entity}, which localizes targets with bounding boxes. Early RES methods adopt a two-stage pipeline \cite{yu2018mattnet, liu2022instance}, where region proposals are generated and then selected via cross-modal matching. Most subsequent methods use a one-stage design, directly fusing visual and linguistic features for end-to-end per-pixel segmentation with FCN-like architectures \cite{cgan, mcn, lts, TIP_RIS_1, TIP_RIS_2}, such as RefSegformer \cite{TIP_RIS_1}.

Recently, owing to the rise of transformers in multi-modal learning, a series of works based on transformers have emerged \cite{VLT,lavt,cris,cgformer,gres}. VLT \cite{VLT} queries the whole image with the generated quires of different attention weights for language expressions. RefTR \cite{reftr} and VG-LAW\cite{vg-law} are the multi-task framework similar to SeqTR \cite{seqtr} and PolyFormer \cite{polyformer}, which models both RES and REC tasks together. Different from RefTR and VG-LAW, SeqTR and PolyFormer formulate the RES as a unified sequential polygon generation task, while RefTR and VG-LAW generate the bounding box and mask via task-specific heads. CRIS \cite{cris} and RISCLIP \cite{RISCLIP} are the CLIP-driven methods that transfer knowledge from CLIP to apply text-to-pixel alignments. CGFormer~\cite{cgformer} uses token grouping with contrastive learning for RES.
MagNet~\cite{magnet} adds a mask grounding task to better align language and vision.

\subsection{Generalized Referring Expression Segmentation}
Recently, \cite{gres} proposed Generalized Referring Expression Segmentation (GRES), a more realistic and challenging extension of classical RES \cite{gsva}. Unlike classical RES, which assumes each expression refers to exactly one target, GRES allows expressions to refer to zero, one, or multiple objects across different categories. This introduces new challenges such as semantic ambiguity, instance grouping, and flexible grounding. For instance, an expression like ``\textit{the animals and the person on the left}" may correspond to multiple heterogeneous objects or none at all.

GRES is still in its early research stage. Representative works include ReLA \cite{gres}, and GSVA \cite{gsva}. ReLA introduces the GRES task and the gRefCOCO dataset, and proposes a baseline model that jointly learns relational localization and segmentation. {
GSVA incorporates MLLMs and SAM, and introduces a special token to support No-target prediction within GRES.}
Unlike existing GRES methods that are specifically designed for the task, our approach focuses on fully leveraging SAM’s capabilities for language-guided segmentation, yet it naturally generalizes to GRES without task-specific designs.

\subsection{SAM-based Grounding}
Segment Anything Model (SAM) \cite{sam} is a powerful universal segmentation model trained on over 1 billion masks. It generates high-quality masks conditioned on geometric prompts such as points and boxes. 
{Most conventional SAM-based RES methods follow a two-stage pipeline and can be classified into two paradigms \cite{SAM_Survey}: Pre-SAM and Post-SAM.}

Pre-SAM approaches generate prompt inputs from text using external models. For example, FGVP \cite{fgvp} and Grounded SAM \cite{grounded_sam} leverage pre-trained vision-language models to produce language-related bounding boxes or point prompts, which are then used to guide SAM. CLIP Surgery \cite{li2023clip_surgery} samples points from attention regions to refine mask proposals. Post-SAM methods treat SAM as a mask proposal generator: TAS \cite{tas} and Semi-SAM \cite{semiSAM} generate a set of candidate masks and use CLIP for cross-modal matching. However, these two-stage pipelines are limited by the quality of the detectors or proposals, and often overlook crucial spatial and relational cues, leading to suboptimal performance on RES.

To address these limitations, beyond the commonly used Pre-SAM and Post-SAM strategies, several recent methods directly leverage SAM’s segmentation features within a one-stage pipeline. For instance, RISAM \cite{RISAM} enhances SAM's multi-scale feature representation through a customized fusion module. In contrast, layer-wise approaches such as DIT \cite{DIT-H} and ReferSAM \cite{tcsvt_refersam} involve extensive architectural modifications to SAM, enabling fine-grained tuning at the cost of increased complexity. Alternatively, prompt-based methods like PPT \cite{PPT} and Prompt-RIS\cite{prompt_ris} preserve SAM’s structure and leverage CLIP to generate geometric prompts (e.g., points, boxes, or masks) that guide SAM’s segmentation. While these approaches reduce dependency on explicit object detectors, they still rely heavily on geometric cues or require intricate training pipelines, limiting their flexibility and generalization.

With the recent advances in Multimodal Large Language Models (MLLMs), a new line of approaches has emerged. Methods such as LISA \cite{LISA}, GSVA \cite{gsva}, RefSAM \cite{refsam}, and LaSagnA \cite{LaSagnA} integrate SAM’s visual tokens into an MLLM and decode segmentation masks through specially designed output tokens. Despite their effectiveness, these methods demand significant computational resources and large-scale data for training and fine-tuning.

Beyond RES, SAM has been applied to tasks such as open-vocabulary recognition \cite{open-vocabulary-sam}, video understanding \cite{sa2va, mpgsam} and zero-shot segmentation \cite{sam_clip}. Inspired by VRP-SAM \cite{vrp_sam}, we propose a lightweight Semantic-Spatial Prompt (SSP) encoder that directly integrates visual and linguistic cues to generate prompts for fully one-stage, language-guided segmentation.

%% file: table/main_results.tex
\begin{table*}[t]
\caption{Comparison with state-of-the-art methods on four referring image segmentation benchmarks. \textit{gIoU} and \textit{cIoU} denote the IoU metrics averaged per instance and accumulated over the entire dataset, respectively. The methods marked with $\ddag$ are trained on the combined RefCOCO/+/g dataset. The methods marked with $\dagger$ are adapted methods. The methods shown in \textcolor{gray}{gray} indicate MLLM-based models trained on large-scale external datasets with much larger text encoders. DN53 refers to DarkNet-53, and RN101 refers to ResNet-101, respectively. The highest performance is marked in bold, and the second-highest performance is marked with an underline.}
\definecolor{lightblue}{RGB}{240, 248, 255} 
\centering
\small
\begin{tabular}{l|cc|ccc|ccc|cc|c} 
\toprule 
\multicolumn{1}{l|}{\multirow{2}{*}{Method}} & \multirow{2}{*}{\begin{tabular}[c]{@{}c@{}}Image\\ Encoder\end{tabular}} & \multirow{2}{*}{\begin{tabular}[c]{@{}c@{}}Text\\ Encoder\end{tabular}} & \multicolumn{3}{c|}{RefCOCO} & \multicolumn{3}{c|}{RefCOCO+} & \multicolumn{2}{c|}{RefCOCOg} & \multicolumn{1}{c}{ReferIt} \\ \cline{4-12} 
                         &          &                                                                                                                                                    & val     & testA   & testB   & val      & testA   & testB   & val  & test & test
                         \\ \hline
\multicolumn{12}{c}{\textit{gIoU}} \\ \hline
\rowcolor{Gray}
\multicolumn{12}{l}{\textbf{Non-SAM based}} \\ 
VLT \cite{seqtr}  &  DN53 &  Bi-GRU & 65.65  & 68.29  & 62.73  & 55.50  & 59.20  & 49.36  & 52.99  & 56.65 & - \\
CRIS~\cite{cris} & RN101    & GPT-2 & 70.47   & 73.18   & 66.10   & 62.27    & 68.08   & 53.68   & 59.87   & 60.36    \\
SeqTR \cite{seqtr} & DN53 & Bi-GRU   & 71.70  & 73.31  & 69.82  & 63.04  & 66.73  & 58.97  & 64.69  & 65.74 & - \\
RefTR~\cite{reftr} & RN101  & BERT & 74.34   & 76.77   & 70.87   & 66.75   & 70.58   & 59.40   & 66.63   & 67.39    \\
PVD~\cite{PVD} & Swin-B & BERT & 74.82 & 77.11 & 69.52 & 63.38 & 68.60 & 56.92 & 63.13 & 63.62 & - \\
LAVT~\cite{lavt} & Swin-B   & BERT & 74.46   & 76.89   & 70.94   & 65.81    & 70.97   & 59.23   & 63.34   & 63.62    \\
VG-LAW \cite{vg-law} & ViT-B & BERT     &   75.62    &    77.51    &    72.89    &    66.63    &    70.38    &    58.89    &    65.63    &  66.08 & - \\
PolyFormer$^\ddag$ \cite{polyformer}  & Swin-B & BERT & 75.96  & 77.09  & 73.22  & 70.65  & 74.51  & \textbf{64.64}  & 69.36  & 69.88 & 65.98   \\
RISCLIP~\cite{RISCLIP}  & CLIP-B & CLIP  & 75.68   & 78.01   & 72.46   & 69.16   & 73.53   & 60.68   & 67.62   & 67.97    \\

DETRIS~\cite{DETRIS} & DINOv2-B & CLIP & 76.0 & 78.2 & 73.5 & 68.9 & 74.0 & 61.5 & 67.9 & 68.1 & - \\
CGFormer~\cite{cgformer} & Swin-B   & BERT & 76.93   & 78.70   & 73.32   & 68.56    & 73.76   & 61.72 & 67.57   & 67.83 & 66.42 \\
\hline
\rowcolor{Gray}
\multicolumn{12}{l}{\textbf{SAM based}} \\ 

SAM$^\dagger$ \cite{sam}  & SAM-H & CLIP  & 57.23   & 58.71  & 56.34  & 44.95  &  50.35  & 41.38  & 47.87  & 49.31  & - \\
RISAM \cite{RISAM}  & SAM-H & CLIP      &  76.20  & 78.92  & 71.84  & 66.37  &  72.10  & 57.33  & 65.48  &  66.60 & - \\
Prompt-RIS~\cite{prompt_ris} & SAM-H & CLIP & \underline{78.10} & \underline{81.21} & \underline{74.64} & \underline{71.13} & \underline{76.60} & \underline{64.25} & \underline{70.47} & \underline{71.29} & - \\ 

\hline
\rowcolor{lightblue}
SSP-SAM-224  & SAM-H & CLIP   &  77.91  & 80.67  & 73.71  & 69.09  &  74.68  & 60.92  & 69.72  &  70.28 & \underline{67.26} \\ 
\rowcolor{lightblue}
SSP-SAM-336 & SAM-H  & CLIP   &  \textbf{79.37}  & \textbf{81.95}  & \textbf{75.52}  & \textbf{71.15}  &  \textbf{77.01}  & 61.78  & \textbf{72.29}  &  \textbf{72.41} & \textbf{67.95} \\
\hline

\multicolumn{12}{c}{\textit{cIoU}} \\ \hline
\rowcolor{Gray}
\multicolumn{12}{l}{\textbf{Non-SAM based}} \\ 
LAVT~\cite{lavt} & Swin-B   & BERT  & 72.73   & 75.82   & 68.79   & 62.14    & 68.38   & 55.10   & 61.24   & 62.09 & -   \\
ReLA~\cite{gres} & Swin-B & BERT & 73.82 & 76.48 & 70.18 & 66.04 & 71.02 & 57.65 & 65.00 & 65.97 & - \\
LQMFormer~\cite{LQMFormer}  & Swin-B   & BERT & 74.16   & 76.82   & 71.04   & 65.91    & 71.84 & 57.59 & 64.73 & 66.04 &-   \\
ReMamber~\cite{remamber}  & Mamba-B   & CLIP & 74.54   & 76.74   & 70.89 & 65.00    & 70.78   & 57.53   & 63.90   & 64.00 &-   \\
CGFormer~\cite{cgformer}  & Swin-B   & BERT & 74.75   & 77.30   & 70.64   & 64.54 & 71.00   & 57.14   & 64.68   & 65.09 & 66.32   \\
PolyFormer$^\ddag$ \cite{polyformer}  & Swin-B & BERT & 74.82 & 76.64 & 71.06 & \underline{67.64} & \underline{72.89} & \underline{59.33} & \underline{67.76} & 69.05 & 71.91   \\

MagNet~\cite{magnet} & Swin-B & BERT & 75.24 & 78.24 & 71.05 & 66.16 & 71.32 & 58.14 & 65.36 & 66.03 & - \\
\hline
\rowcolor{Gray}
\multicolumn{12}{l}{\textbf{SAM based}} \\ 
EVF-SAM-CLIP~\cite{evf-sam} & SAM-H & CLIP & 61.0 & 63.4 & 59.9 & 43.1 & 45.9 & 40.6 & 48.9 & 49.6 & - \\
\textit{Layer-wise} DIT-B~\cite{DIT-H} & SAM-B & BERT & 71.98 & 74.51 & 68.77 & 59.97 & 65.52 & 51.72 & 60.18 & 61.15 & - \\
\textit{Layer-wise} DIT-H~\cite{DIT-H} & SAM-H & BERT & \underline{76.18} & 78.13 & \textbf{73.27} & \textbf{68.00} & 71.77 & \textbf{60.04} & 67.37 & \underline{67.92} & - \\
\textcolor{gray!60}{VideoLISA-3.8B~\cite{videoLISA}} & \textcolor{gray!60}{SAM-H} & \textcolor{gray!60}{Phi-3-3.8B} & \textcolor{gray!60}{73.8} & \textcolor{gray!60}{76.6} & \textcolor{gray!60}{68.8} & \textcolor{gray!60}{63.4} & \textcolor{gray!60}{68.8} & \textcolor{gray!60}{56.2} & \textcolor{gray!60}{68.3} & \textcolor{gray!60}{68.8} & \textcolor{gray!60}{-} \\
\textcolor{gray!60}{RefSAM~\cite{refsam}} & \textcolor{gray!60}{SAM-H} & \textcolor{gray!60}{T5-3B} & \textcolor{gray!60}{74.89} & \textcolor{gray!60}{77.49} & \textcolor{gray!60}{70.90} & \textcolor{gray!60}{65.88} & \textcolor{gray!60}{71.60} & \textcolor{gray!60}{57.26} & \textcolor{gray!60}{66.46} & \textcolor{gray!60}{67.23} & \textcolor{gray!60}{-} \\
\textcolor{gray!60}{LISA-Vicuna-7B~\cite{LISA}} & \textcolor{gray!60}{SAM-H} & \textcolor{gray!60}{Vicuna-7B} & \textcolor{gray!60}{74.9}  & \textcolor{gray!60}{79.1}  &  \textcolor{gray!60}{72.3}  & \textcolor{gray!60}{65.1}  &  \textcolor{gray!60}{70.8}  & \textcolor{gray!60}{58.1}  & \textcolor{gray!60}{67.9}  & \textcolor{gray!60}{70.6} & \textcolor{gray!60}{-}   \\ 
\textcolor{gray!60}{GSVA-Vicuna-7B~\cite{gsva}} & \textcolor{gray!60}{SAM-H} & \textcolor{gray!60}{Vicuna-7B} & \textcolor{gray!60}{76.4} & \textcolor{gray!60}{77.4} & \textcolor{gray!60}{72.8} & \textcolor{gray!60}{64.5} & \textcolor{gray!60}{67.7} & \textcolor{gray!60}{58.6} & \textcolor{gray!60}{71.1} & \textcolor{gray!60}{72.0} & \textcolor{gray!60}{-}\\
\textcolor{gray!60}{LaSagnA-7B~\cite{LaSagnA}} & \textcolor{gray!60}{SAM-H} & \textcolor{gray!60}{Vicuna-7B} & \textcolor{gray!60}{76.8} & \textcolor{gray!60}{78.7} & \textcolor{gray!60}{\textbf{73.8}} & \textcolor{gray!60}{64.4} & \textcolor{gray!60}{70.6} & \textcolor{gray!60}{60.1} & \textcolor{gray!60}{70.6} & \textcolor{gray!60}{71.9} & \textcolor{gray!60}{-} \\

\hline
\rowcolor{lightblue}SSP-SAM-224 & SAM-H & CLIP & 75.12 & \underline{78.58} & 70.64  & 63.87 & 70.60 & 54.41 & 65.32 & 66.15 & \underline{72.65} \\
\rowcolor{lightblue}SSP-SAM-336 & SAM-H & CLIP & \textbf{77.12} & \textbf{80.04} & \underline{72.41} & 66.03 & \textbf{73.45} & 55.29  & \textbf{68.25} & \textbf{68.08} & \textbf{73.04} \\
\bottomrule

\end{tabular}

\label{tab:res_sota}

\end{table*}

%% file: table/main_results_pr.tex
\begin{table}[t]
\caption{Comparisons of Pr@X results on the RefCOCO val split under classic RES settings. The best performance is highlighted in bold, and the second-best is underlined. Our method tends to generate higher-precision masks, as evidenced by its superior performance under the stricter Pr@0.9 metric.}
\definecolor{lightblue}{RGB}{240, 248, 255} 
  \centering
  \setlength{\tabcolsep}{2mm}{
    \resizebox{\linewidth}{!}{
    \begin{tabular}{lccccc}
    \toprule
    \multirow{2}{*}{Method} & \multicolumn{5}{c}{RefCOCO val} \\
      & Pr@0.9 & Pr@0.7 & Pr@0.5 & gIoU & cIoU \\
    \midrule
    LAVT \cite{lavt} & 34.30 & 75.28  & 84.46 & 74.46 & 72.73 \\
    ReLA \cite{gres} & 34.99 & 77.71 & 85.82 & 75.61 & 73.82 \\
    CGFormer \cite{cgformer} & 38.77 & 78.69 & 87.23 & 76.93 & 74.75 \\ 
    MagNet \cite{magnet} & 38.77 & 80.01 & \underline{87.79} & 77.43 & \underline{75.24} \\
    \midrule
    \rowcolor{lightblue}SSP-SAM-224 & \underline{46.11} & \underline{81.42} & 87.18 & \underline{77.91} & 75.12 \\
    \rowcolor{lightblue}SSP-SAM-336 & \textbf{47.77} & \textbf{83.15} & \textbf{88.80} & \textbf{79.37} & \textbf{77.12} \\
    \bottomrule
  \end{tabular}
  \label{tab:res_pr}
  }
}
\end{table}

%% file: table/gres.tex
\begin{table*}
\definecolor{lightblue}{RGB}{240, 248, 255} 
\caption{Generalized referring image segmentation (GRES) results on gRefCOCO dataset. \textit{gIoU} and \textit{cIoU} denote the IoU metrics averaged per instance and accumulated over the entire dataset, respectively. N-acc. is short for the No-target accuracy. (ft) denotes the model is finetuned on the training set of gRefCOCO.}
\begin{center}
\setlength{\tabcolsep}{3.5mm}{
\renewcommand\arraystretch{1.0}
\resizebox{\linewidth}{!}{
\begin{tabular}{l|ccc|ccc|ccc}
\toprule
\multirow{2}{*}{Method} & \multicolumn{3}{c|}{val} & \multicolumn{3}{c|}{testA}  & \multicolumn{3}{c}{testB} \\
& gIoU & cIoU & N-acc. & gIoU & cIoU & N-acc. & gIoU & cIoU & N-acc. \\
\hline
\rowcolor{Gray}
\multicolumn{10}{l}{\textbf{Non-SAM based}} \\ 
MattNet~\cite{yu2018mattnet} & 48.24 & 47.51 & 41.15 & 59.30 & 58.66 & 44.04 & 46.14 & 45.33 & 41.32 \\
LTS~\cite{lts} & 52.70 & 52.30 & - & 62.64 & 61.87 & - & 50.42 & 49.96 & - \\
VLT~\cite{VLT} & 52.00 & 52.51 & 47.17 & 63.20 & 62.19 & 48.74 & 50.88 & 50.52 & 47.82 \\
CRIS~\cite{cris} & 56.27 & 55.34 & - & 63.42 & 63.82 & - & 51.79 & 51.04  & - \\
LAVT~\cite{lavt} & 58.40 & 57.64 & 49.32 & 65.90 & 65.32 & 49.25 & 55.83 & 55.04 & 48.46 \\
CGFormer~\cite{cgformer} & 63.01 & 62.28 & - & 70.13 & 68.15 & - & 61.09 & 60.18 & - \\
ReLA~\cite{gres} & 63.60 & 62.42 & 56.37 & 70.03 & 69.26 & 59.02 & 61.02 & 59.88 & 58.40 \\

\hline
\rowcolor{Gray}
\multicolumn{10}{l}{\textbf{SAM based}} \\ 
LISA-Vicuna-7B~\cite{LISA} & 32.21 & 38.72 & 2.71 & 48.54 & 52.55 & 6.37 & 39.65 & 44.79 & 5.00 \\
GSVA-Vicuna-7B~\cite{gsva} & 63.32 & 61.70 & 56.45 & 70.11 & 69.23 & 63.50 & 61.34 & 60.26 & 58.42  \\
LISA-Vicuna-7B (ft)~\cite{LISA} & 61.63 & 61.76 & 54.67 & 66.27 & 68.50 & 50.01 & 58.84 & 60.63 & 51.91 \\
GSVA-Vicuna-7B (ft)~\cite{gsva} & 66.47 & \textbf{63.29} & 62.43 & 71.08 & \underline{69.93} & \underline{65.31} & \underline{62.23} & 60.47 & \underline{60.56} \\
\hline
\rowcolor{lightblue}SSP-SAM-224 & \underline{68.87} & 61.28 & \textbf{69.37} & \underline{71.79} & 68.97 & \textbf{69.23} & 61.55 & \underline{58.24}  & \textbf{61.36} \\
\rowcolor{lightblue}SSP-SAM-336 & \textbf{69.09} & \underline{62.56} & \underline{66.12} & \textbf{72.79} & \textbf{70.86} & 65.08 & \textbf{62.81} & \textbf{60.91} & 57.19 \\

\bottomrule
\end{tabular}}
}
\end{center}
\label{tab:gres}
\end{table*}

%% file: table/gres_pr.tex
\begin{table}[t]
\caption{Comparisons of Pr@X results on the gRefCOCO validation split under the GRES settings. The best performance is highlighted in bold, and the second-best is underlined. Our method consistently tends to produce higher-precision masks, as reflected by its strong performance under the strict Pr@0.9 metric.}
\definecolor{lightblue}{RGB}{240, 248, 255} 
  \centering
  \setlength{\tabcolsep}{2mm}{
    \resizebox{\linewidth}{!}{
    \begin{tabular}{lccccc}
    \toprule
    \multirow{2}{*}{Method} & \multicolumn{5}{c}{gRefCOCO val} \\ 
    
      & Pr@0.9 & Pr@0.8 & Pr@0.7 & gIoU & cIoU \\
    \midrule
    CGFormer \cite{cgformer} & 22.43 & 56.57  & \underline{68.93} & 63.01 & 62.28 \\
    ReLA \cite{gres} & 24.68  & \textbf{68.33} & \textbf{74.20}  & 63.60 & \underline{62.42}  \\
    \midrule
    \rowcolor{lightblue}SSP-SAM-224 & \underline{44.00} & 60.55 & 66.49 & \underline{68.87} & 61.28 \\
    \rowcolor{lightblue}SSP-SAM-336 & \textbf{44.77}  & \underline{61.54}  & 67.33 & \textbf{69.09} & \textbf{62.56} \\
    \bottomrule
  \end{tabular}
  \label{tab:gres_pr}
  }
}
\end{table}

%% file: table/phrase_cut.tex
\begin{table}[t]
\caption{Comparisons in the open-vocabulary scenario on the PhraseCut test split. The best performance is highlighted in bold, and the second-best is underlined. The evaluation metric is the default \textit{gIoU}. Our method demonstrates improved generalization compared to other approaches.}
\definecolor{lightblue}{RGB}{240, 248, 255} 
\centering
\setlength{\tabcolsep}{0.8mm}
\resizebox{\linewidth}{!}{
\begin{tabular}{lccccc}
\toprule
\multirow{2}{*}{Training Set} & \multicolumn{5}{c}{PhraseCut test} \\ 
 & CRIS & LAVT & RISAM & \cellcolor{lightblue}SSP-SAM-224 & \cellcolor{lightblue}SSP-SAM-336 \\
\midrule
RefCOCO & 15.53 & 16.68 & 22.82 & \cellcolor{lightblue}\underline{24.66} & \cellcolor{lightblue}\textbf{24.69} \\
RefCOCO+ & 16.30 & 16.64 & 21.68 & \cellcolor{lightblue}\textbf{25.75} & \cellcolor{lightblue}\underline{24.91} \\
RefCOCOg & 16.24 & 16.05 & 22.47 & \cellcolor{lightblue}\underline{27.00} & \cellcolor{lightblue}\textbf{28.29} \\ 
\bottomrule
\end{tabular}
}
\label{tab:zero-shot}
\end{table}

%% file: refer.bib
@String(AAAI = {AAAI})

@article{li2023clip_surgery,
  author       = {Yi Li and
                  Hualiang Wang and
                  Yiqun Duan and
                  Jiheng Zhang and
                  Xiaomeng Li},
  title        = {A closer look at the explainability of Contrastive language-image pre-training},
  journal      = {Pattern Recognition},
  volume       = {162},
  pages        = {111409},
  year         = {2025},
}

@article{liu2022instance,
  title={Instance-specific feature propagation for referring segmentation},
  author={Liu, Chang and Jiang, Xudong and Ding, Henghui},
  journal      = {IEEE Trans. Multimedia},
  volume       = {25},
  pages        = {3657--3667},
  year         = {2023},
}

@inproceedings{yu2018mattnet,
  title = {Mattnet: Modular attention network for referring expression comprehension},
  author = {Yu, Licheng and Lin, Zhe and Shen, Xiaohui and Yang, Jimei and Lu, Xin and Bansal, Mohit and Berg, Tamara L},
  booktitle    = {Proceedings of IEEE/CVF Conference on Computer Vision and Pattern Recognition},
  pages = {1307--1315},
  year = {2018}
}

@inproceedings{VLT,
  title={Vision-language transformer and query generation for referring segmentation},
  author={Ding, Henghui and Liu, Chang and Wang, Suchen and Jiang, Xudong},
  booktitle    = {Proceedings of IEEE/CVF International Conference on Computer Vision},
  pages={16321--16330},
  year={2021}
}

@inproceedings{tas,
  title={Text augmented spatial-aware zero-shot referring image segmentation},
  author={Suo, Yucheng and Zhu, Linchao and Yang, Yi},
  booktitle    = {Findings of the Association for Computational Linguistics},
  pages        = {1032--1043},
  year={2023}
}

@inproceedings{fgvp,
  title={Fine-grained visual prompting},
  author={Yang, Lingfeng and Wang, Yueze and Li, Xiang and Wang, Xinlong and Yang, Jian},
  booktitle= {Proceedings of Advances in Neural Information Processing Systems},
  volume={36},
  year={2024}
}

@inproceedings{cgan,
  title={Cascade grouped attention network for referring expression segmentation},
  author={Luo, Gen and Zhou, Yiyi and Ji, Rongrong and Sun, Xiaoshuai and Su, Jinsong and Lin, Chia-Wen and Tian, Qi},
  booktitle={Proceedings of ACM International Conference on Multimedia},
  pages={1274--1282},
  year={2020}
}

@inproceedings{mcn,
  title={Multi-task collaborative network for joint referring expression comprehension and segmentation},
  author={Luo, Gen and Zhou, Yiyi and Sun, Xiaoshuai and Cao, Liujuan and Wu, Chenglin and Deng, Cheng and Ji, Rongrong},
  booktitle    = {Proceedings of IEEE/CVF Conference on Computer Vision and Pattern Recognition},
  pages={10034--10043},
  year={2020}
}

@inproceedings{lts,
  title={Locate then segment: A strong pipeline for referring image segmentation},
  author={Jing, Ya and Kong, Tao and Wang, Wei and Wang, Liang and Li, Lei and Tan, Tieniu},
  booktitle    = {Proceedings of IEEE/CVF Conference on Computer Vision and Pattern Recognition},
  pages={9858--9867},
  year={2021}
}

@inproceedings{reftr,
  author       = {Muchen Li and
                  Leonid Sigal},
  title        = {Referring transformer: {A} one-step approach to multi-task visual grounding},
    booktitle    = {Proceedings of Advances in Neural Information Processing Systems},
  pages        = {19652--19664},
  year         = {2021},
}

@inproceedings{seqtr,
  author    = {Chaoyang Zhu and
               Yiyi Zhou and
               Yunhang Shen and
               Gen Luo and
               Xingjia Pan and
               Mingbao Lin and
               Chao Chen and
               Liujuan Cao and
               Xiaoshuai Sun and
               Rongrong Ji},
  title     = {SeqTR: {A} simple Yet universal network for visual grounding},
  booktitle={Proceedings of European Conference on Computer Vision},
  volume    = {13695},
  pages     = {598--615},
  year      = {2022}
}

@inproceedings{polyformer,
  title={Polyformer: Referring image segmentation as sequential polygon generation},
  author={Liu, Jiang and Ding, Hui and Cai, Zhaowei and Zhang, Yuting and Satzoda, Ravi Kumar and Mahadevan, Vijay and Manmatha, R},
  booktitle    = {Proceedings of IEEE/CVF Conference on Computer Vision and Pattern Recognition},
  pages={18653--18663},
  year={2023}
}

@inproceedings{cris,
  title={Cris: Clip-driven referring image segmentation},
  author={Wang, Zhaoqing and Lu, Yu and Li, Qiang and Tao, Xunqiang and Guo, Yandong and Gong, Mingming and Liu, Tongliang},
  booktitle    = {Proceedings of IEEE/CVF Conference on Computer Vision and Pattern Recognition},
  pages={11686--11695},
  year={2022}
}

@inproceedings{lavt,
  title={Lavt: Language-aware vision transformer for referring image segmentation},
  author={Yang, Zhao and Wang, Jiaqi and Tang, Yansong and Chen, Kai and Zhao, Hengshuang and Torr, Philip HS},
  booktitle    = {Proceedings of IEEE/CVF Conference on Computer Vision and Pattern Recognition},
  pages={18155--18165},
  year={2022}
}

@inproceedings{cgformer,
  title={Contrastive grouping with transformer for referring image segmentation},
  author={Tang, Jiajin and Zheng, Ge and Shi, Cheng and Yang, Sibei},
  booktitle    = {Proceedings of IEEE/CVF Conference on Computer Vision and Pattern Recognition},
  pages={23570--23580},
  year={2023}
}

@article{RISAM,
  title={RISAM: Referring image segmentation via mutual-aware attention features},
  author={Zhang, Mengxi and Liu, Yiming and Yin, Xiangjun and Yue, Huanjing and Yang, Jingyu},
  journal={arXiv preprint arXiv:2311.15727},
  year={2023}
}

@article{refsam,
  title={Refsam: Efficiently adapting segmenting anything model for referring video object segmentation},
  author={Li, Yonglin and Zhang, Jing and Teng, Xiao and Lan, Long},
  journal={arXiv preprint arXiv:2307.00997},
  year={2023}
}

@inproceedings{vg-law,
  title={Language adaptive weight generation for multi-task visual grounding},
  author={Su, Wei and Miao, Peihan and Dou, Huanzhang and Wang, Gaoang and Qiao, Liang and Li, Zheyang and Li, Xi},
  booktitle    = {Proceedings of IEEE/CVF Conference on Computer Vision and Pattern Recognition},
  pages={10857--10866},
  year={2023}
}

@inproceedings{sam,
  title={Segment anything},
  author={Kirillov, Alexander and Mintun, Eric and Ravi, Nikhila and Mao, Hanzi and Rolland, Chloe and Gustafson, Laura and Xiao, Tete and Whitehead, Spencer and Berg, Alexander C and Lo, Wan-Yen and others},
  booktitle    = {Proceedings of IEEE/CVF International Conference on Computer Vision},
  pages={4015--4026},
  year={2023}
}

@article{grounded_sam,
  title={Grounded sam: Assembling open-world models for diverse visual tasks},
  author={Ren, Tianhe and Liu, Shilong and Zeng, Ailing and Lin, Jing and Li, Kunchang and Cao, He and Chen, Jiayu and Huang, Xinyu and Chen, Yukang and Yan, Feng and others},
  journal={arXiv preprint arXiv:2401.14159},
  year={2024}
}

@inproceedings{sam_clip,
  author       = {Haoxiang Wang and
                  Pavan Kumar Anasosalu Vasu and
                  Fartash Faghri and
                  Raviteja Vemulapalli and
                  Mehrdad Farajtabar and
                  Sachin Mehta and
                  Mohammad Rastegari and
                  Oncel Tuzel and
                  Hadi Pouransari},
  title        = {{SAM-CLIP:} Merging vision foundation models towards semantic and
                  spatial Understanding},
  booktitle    = {Proceedings of IEEE/CVF International Conference on Computer Vision},
  pages        = {3635--3647},
  year         = {2024},
}

@inproceedings{vrp_sam,
  title={VRP-SAM: SAM with visual reference prompt},
  author={Sun, Yanpeng and Chen, Jiahui and Zhang, Shan and Zhang, Xinyu and Chen, Qiang and Zhang, Gang and Ding, Errui and Wang, Jingdong and Li, Zechao},
  booktitle={Proceedings of IEEE/CVF Conference on Computer Vision and Pattern Recognition},
  year={2024}
}

@inproceedings{clip,
  title={Learning transferable visual models from natural language supervision},
  author={Radford, Alec and Kim, Jong Wook and Hallacy, Chris and Ramesh, Aditya and Goh, Gabriel and Agarwal, Sandhini and Sastry, Girish and Askell, Amanda and Mishkin, Pamela and Clark, Jack and others},
  booktitle={Proceedings of International Conference on Machine Learning},
  pages={8748--8763},
  year={2021},
}

@inproceedings{referit,
  title={Referitgame: Referring to objects in Photographs of Natural Scenes},
  author={Kazemzadeh, Sahar and Ordonez, Vicente and Matten, Mark and Berg, Tamara},
  booktitle={Proceedings of Conference on Empirical Methods in Natural Language Processing},
  pages={787--798},
  year={2014}
}

@inproceedings{refcoco_umd,
  title={Modeling context between Objects for referring expression Understanding},
  author={Nagaraja, Varun K and Morariu, Vlad I and Davis, Larry S},
  booktitle={Proceedings of European Conference on Computer Vision},
  pages={792--807},
  year={2016}
}

@inproceedings{refcocog_google,
  title={Generation and comprehension of unambiguous object descriptions},
  author={Mao, Junhua and Huang, Jonathan and Toshev, Alexander and Camburu, Oana and Yuille, Alan L and Murphy, Kevin},
  booktitle={Proceedings of IEEE/CVF International Conference on Computer Vision},
  pages={11--20},
  year={2016}
}

@inproceedings{mscoco,
  title={Microsoft COCO: Common objects in context},
  author={Lin, Tsung-Yi and Maire, Michael and Belongie, Serge and Hays, James and Perona, Pietro and Ramanan, Deva and Doll{\'a}r, Piotr and Zitnick, C Lawrence},
  booktitle={Proceedings of European Conference on Computer Vision},
  pages={740--755},
  year={2014}
}

@inproceedings{vltvg,
  title={Improving visual grounding with visual-linguistic verification and iterative reasoning},
  author={Yang, Li and Xu, Yan and Yuan, Chunfeng and Liu, Wei and Li, Bing and Hu, Weiming},
  booktitle={Proceedings of IEEE/CVF Conference on Computer Vision and Pattern Recognition},
  pages={9499--9508},
  year={2022}
}

@article{transcp,
  author    = {Tang, Wei and
               Li, Liang and
               Liu, Xuejing and
               Jin, Lu and
               Tang, Jinhui and
               Li, Zechao
              },
  title     = {Context disentangling and prototype inheriting for robust visual grounding},
  journal   = {IEEE Trans. Pattern Anal. Mach. Intell.},
  volume       = {46},
  number       = {5},
  pages        = {3213--3229},
  year         = {2024},
}

@inproceedings{transvg,
  title={Transvg: end-to-end visual grounding with transformers},
  author={Deng, Jiajun and Yang, Zhengyuan and Chen, Tianlang and Zhou, Wengang and Li, Houqiang},
  booktitle={Proceedings of IEEE/CVF International Conference on Computer Vision},
  pages={1769--1779},
  year={2021}
}

@article{flickr30k,
  author    = {Bryan A. Plummer and
               Liwei Wang and
               Chris M. Cervantes and
               Juan C. Caicedo and
               Julia Hockenmaier and
               Svetlana Lazebnik},
  title     = {Flickr30k Entities: Collecting region-to-phrase correspondences for richer image-to-sentence models},
  journal   = {Int. J. Comput. Vis.},
  volume    = {123},
  number    = {1},
  pages     = {74--93},
  year      = {2017}
}

@article{visual_genome,
  author       = {Ranjay Krishna and
                  Yuke Zhu and
                  Oliver Groth and
                  Justin Johnson and
                  Kenji Hata and
                  Joshua Kravitz and
                  Stephanie Chen and
                  Yannis Kalantidis and
                  Li{-}Jia Li and
                  David A. Shamma and
                  Michael S. Bernstein and
                  Li Fei{-}Fei},
  title        = {Visual Genome: connecting language and vision Using crowdsourced dense
                  image annotations},
  journal   = {Int. J. Comput. Vis.},
  volume       = {123},
  number       = {1},
  pages        = {32--73},
  year         = {2017},
}

@article{liu2022entity,
  author={Liu, Xuejing and Li, Liang and Wang, Shuhui and Zha, Zheng-Jun and Li, Zechao and Tian, Qi and Huang, Qingming},
  journal={IEEE Trans. Pattern Anal. Mach. Intell.}, 
  title={Entity-enhanced adaptive reconstruction network for weakly supervised referring expression grounding}, 
  year={2023},
  volume={45},
  number={3},
  pages={3003-3018}
  }

@article{ctnet,
  author       = {Zechao Li and
                  Yanpeng Sun and
                  Liyan Zhang and
                  Jinhui Tang},
  title        = {CTNet: Context-based tandem network for semantic segmentation},
  journal      = {IEEE Trans. Pattern Anal. Mach. Intell.},
  volume       = {44},
  number       = {12},
  pages        = {9904--9917},
  year         = {2022},
}

@inproceedings{open-vocabulary-sam,
  author       = {Haobo Yuan and
                  Xiangtai Li and
                  Chong Zhou and
                  Yining Li and
                  Kai Chen and
                  Chen Change Loy},
  title        = {Open-Vocabulary {SAM:} Segment and recognize twenty-thousand classes
                  interactively},
  booktitle    = {Proceedings of European Conference on Computer Vision},
  volume       = {15101},
  pages        = {419--437},
  year         = {2024},
}

@inproceedings{semiSAM,
  author       = {Danni Yang and
                  Jiayi Ji and
                  Yiwei Ma and
                  Tianyu Guo and
                  Haowei Wang and
                  Xiaoshuai Sun and
                  Rongrong Ji},
  title        = {{SAM} as the Guide: Mastering pseudo-label refinement in semi-supervised
                  referring expression segmentation},
  booktitle    = {Proceedings of International Conference on Machine Learning},
  year         = {2024},
}

@article{SAM_Survey,
  title={A comprehensive survey on segment anything model for vision and beyond},
  author={Zhang, Chunhui and Liu, Li and Cui, Yawen and Huang, Guanjie and Lin, Weilin and Yang, Yiqian and Hu, Yuehong},
  journal={arXiv preprint arXiv:2305.08196},
  year={2023}
}

@inproceedings{PPT,
  title={Curriculum point prompting for weakly-supervised referring image segmentation},
  author={Dai, Qiyuan and Yang, Sibei},
  booktitle={Proceedings of IEEE/CVF Conference on Computer Vision and Pattern Recognition},
  pages={13711--13722},
  year={2024}
}

@inproceedings{phrasecut,
  title={Phrasecut: Language-based image segmentation in the wild},
  author={Wu, Chenyun and Lin, Zhe and Cohen, Scott and Bui, Trung and Maji, Subhransu},
  booktitle={Proceedings of IEEE/CVF Conference on Computer Vision and Pattern Recognition},
  pages={10216--10225},
  year={2020}
}

@inproceedings{seg_robot_1,
  author       = {Mohit Shridhar and
                  David Hsu},
  title        = {Interactive visual grounding of referring expressions for human-robot
                  interaction},
  booktitle    = {Proceedings of Robotics: Science and System},
  year         = {2018},
}

@article{TIP_RIS_1,
  author={Wu, Jianzong and Li, Xiangtai and Li, Xia and Ding, Henghui and Tong, Yunhai and Tao, Dacheng},
  journal={IEEE Trans. Image Process.}, 
  title={Toward robust referring image segmentation}, 
  volume       = {33},
  pages        = {1782--1794},
  year         = {2024},
}

@article{TIP_RIS_2,
  title={Multi-modal mutual attention and iterative interaction for referring image segmentation},
  author={Liu, Chang and Ding, Henghui and Zhang, Yulun and Jiang, Xudong},
  journal={IEEE Trans. Image Process.},
  volume       = {32},
  pages        = {3054--3065},
  year         = {2023},
}

@inproceedings{IMGEDIT_RIS,
  title={Referring Image Editing: Object-level Image Editing via Referring Expressions},
  author={Liu, Chang and Li, Xiangtai and Ding, Henghui},
  booktitle={Proceedings of IEEE/CVF Conference on Computer Vision and Pattern Recognition},
  pages={13128--13138},
  year={2024}
}

@inproceedings{magnet,
  title={Mask grounding for referring image segmentation},
  author={Chng, Yong Xien and Zheng, Henry and Han, Yizeng and Qiu, Xuchong and Huang, Gao},
  booktitle={Proceedings of IEEE/CVF Conference on Computer Vision and Pattern Recognition},
  pages={26573--26583},
  year={2024}
}

@inproceedings{PVD,
  title={Parallel vertex diffusion for unified visual grounding},
  author={Cheng, Zesen and Li, Kehan and Jin, Peng and Li, Siheng and Ji, Xiangyang and Yuan, Li and Liu, Chang and Chen, Jie},
  booktitle={Proceedings of AAAI Conference on Artificial Intelligence},
  volume={38},
  number={2},
  pages={1326--1334},
  year={2024}
}

@inproceedings{RISCLIP,
  author       = {Seoyeon Kim and
                  Minguk Kang and
                  Dongwon Kim and
                  Jaesik Park and
                  Suha Kwak},
  title        = {Extending CLIP's image-text alignment to referring image segmentation},
  booktitle    = {Proceedings of the North American Chapter of the Association for Computational Linguistics},
  pages        = {4611--4628},
  year         = {2024},
}

@inproceedings{LQMFormer,
  title={LQMFormer: Language-aware query mask transformer for referring image segmentation},
  author={Shah, Nisarg A and VS, Vibashan and Patel, Vishal M},
  booktitle={Proceedings of IEEE/CVF Conference on Computer Vision and Pattern Recognition},
  pages={12903--12913},
  year={2024}
}

@inproceedings{DETRIS,
  title={Densely connected parameter-efficient tuning for referring image segmentation},
  author={Huang, Jiaqi and Xu, Zunnan and Liu, Ting and Liu, Yong and Han, Haonan and Yuan, Kehong and Li, Xiu},
  booktitle={Proceedings of AAAI Conference on Artificial Intelligence},
  volume={39},
  number={4},
  pages={3653-3661},
  year={2025}
}

@inproceedings{prompt_ris,
  title={Prompt-driven referring image segmentation with instance contrasting},
  author={Shang, Chao and Song, Zichen and Qiu, Heqian and Wang, Lanxiao and Meng, Fanman and Li, Hongliang},
  booktitle={Proceedings of IEEE/CVF Conference on Computer Vision and Pattern Recognition},
  pages={4124--4134},
  year={2024}
}

@inproceedings{remamber,
  title={Remamber: Referring image segmentation with mamba twister},
  author={Yang, Yuhuan and Ma, Chaofan and Yao, Jiangchao and Zhong, Zhun and Zhang, Ya and Wang, Yanfeng},
  booktitle={Proceedings of European Conference on Computer Vision},
  pages={108--126},
  year={2024},
}

@inproceedings{DIT-H,
  title={Deep instruction tuning for segment anything model},
  author={Huang, Xiaorui and Luo, Gen and Zhu, Chaoyang and Tong, Bo and Zhou, Yiyi and Sun, Xiaoshuai and Ji, Rongrong},
  booktitle={Proceedings of ACM International Conference on Multimedia},
  pages={905--914},
  year={2024}
}

@inproceedings{videoLISA,
  title={One token to seg them all: Language instructed reasoning segmentation in videos},
  author={Bai, Zechen and He, Tong and Mei, Haiyang and Wang, Pichao and Gao, Ziteng and Chen, Joya and Zhang, Zheng and Shou, Mike Zheng},
  booktitle={Proceedings of Advances in Neural Information Processing Systems},
  volume={37},
  pages={6833--6859},
  year={2024}
}

@inproceedings{LISA,
  title={Lisa: Reasoning segmentation via large language model},
  author={Lai, Xin and Tian, Zhuotao and Chen, Yukang and Li, Yanwei and Yuan, Yuhui and Liu, Shu and Jia, Jiaya},
  booktitle={Proceedings of IEEE/CVF Conference on Computer Vision and Pattern Recognition},
  pages={9579--9589},
  year={2024}
}

@inproceedings{gsva,
  title={Gsva: Generalized segmentation via multimodal large language models},
  author={Xia, Zhuofan and Han, Dongchen and Han, Yizeng and Pan, Xuran and Song, Shiji and Huang, Gao},
  booktitle={Proceedings of IEEE/CVF Conference on Computer Vision and Pattern Recognition},
  pages={3858--3869},
  year={2024}
}

@article{LaSagnA,
  title={Lasagna: Language-based segmentation assistant for complex queries},
  author={Wei, Cong and Tan, Haoxian and Zhong, Yujie and Yang, Yujiu and Ma, Lin},
  journal={arXiv preprint arXiv:2404.08506},
  year={2024}
}

@inproceedings{gres,
  title={Gres: Generalized referring expression segmentation},
  author={Liu, Chang and Ding, Henghui and Jiang, Xudong},
  booktitle={Proceedings of IEEE/CVF conference on Computer Vision and Pattern Recognition},
  pages={23592--23601},
  year={2023}
}

@ARTICLE{visual_incontext,
  author={Sun, Yanpeng and Chen, Qiang and Wang, Jian and Wang, Jingdong and Li, Zechao},
  journal={IEEE Trans. Image Process.}, 
  title={Exploring effective factors for improving visual in-context learning}, 
  year={2025},
  volume={34},
  number={},
  pages={2147-2160},
  }

@article{vpp_llava,
  title={Visual position prompt for MLLM based visual grounding},
  author={Wei Tang and Yanpeng Sun and Qinying Gu and Zechao Li},
  journal      = {IEEE Trans. Multimedia},
  volume       = {},
  pages        = {},
  year         = {2025},
note={to appear},
}

@ARTICLE{tcsvt_refersam,
  author={Liu, Sun-Ao and Xie, Hongtao and Ge, Jiannan and Zhang, Yongdong},
  journal={IEEE Trans. Circuits Syst. Video Technol.}, 
  title={ReferSAM: Unleashing segment anything model for referring image segmentation}, 
  year={2025},
  volume={35},
  number={5},
  pages={4910-4922},
}

@ARTICLE{tcsvt_crscnet,
  author={Shang, Chao and Li, Hongliang and Qiu, Heqian and Wu, Qingbo and Meng, Fanman and Zhao, Taijin and Ngan, King Ngi},
  journal={IEEE Trans. Circuits Syst. Video Technol.}, 
  title={Cross-modal recurrent semantic comprehension for referring image segmentation}, 
  year={2023},
  volume={33},
  number={7},
  pages={3229-3242}
}

@ARTICLE{tcsvt_bms,
  author={Li, Wenhui and Pang, Chao and Nie, Weizhi and Tian, Hongshuo and Liu, An-An},
  journal={IEEE Trans. Circuits Syst. Video Technol.}, 
  title={Bidirectional mask selection for zero-shot referring image segmentation}, 
  year={2025},
  volume={35},
  number={1},
  pages={911-921},
}

@ARTICLE{tcsvt_pks,
  author={Li, Hui and Sun, Mingjie and Xiao, Jimin and Lim, Eng Gee and Zhao, Yao},
  journal={IEEE Trans. Circuits Syst. Video Technol.}, 
  title={Fully and weakly supervised referring expression segmentation with end-to-end learning}, 
  year={2023},
  volume={33},
  number={10},
  pages={5999-6012}
}

@article{eva-clip,
  title={Eva-clip: Improved training techniques for clip at scale},
  author={Sun, Quan and Fang, Yuxin and Wu, Ledell and Wang, Xinlong and Cao, Yue},
  journal={arXiv preprint arXiv:2303.15389},
  year={2023}
}

@article{siglip2,
  title={Siglip 2: Multilingual vision-language encoders with improved semantic understanding, localization, and dense features},
  author={Tschannen, Michael and Gritsenko, Alexey and Wang, Xiao and Naeem, Muhammad Ferjad and Alabdulmohsin, Ibrahim and Parthasarathy, Nikhil and Evans, Talfan and Beyer, Lucas and Xia, Ye and Mustafa, Basil and others},
  journal={arXiv preprint arXiv:2502.14786},
  year={2025}
}

@article{evf-sam,
      title={EVF-SAM: Early vision-language fusion for text-prompted segment anything model}, 
      author={Yuxuan Zhang and Tianheng Cheng and Rui Hu and Lei Liu and Heng Liu and Longjin Ran and Xiaoxin Chen and Wenyu Liu and Xinggang Wang},
      journal={arXiv preprint arXiv:2406.20076},
      year={2024}
}

@article{gru,
  author       = {Junyoung Chung and
                  {\c{C}}aglar G{\"{u}}l{\c{c}}ehre and
                  KyungHyun Cho and
                  Yoshua Bengio},
  title        = {Empirical Evaluation of Gated Recurrent Neural Networks on Sequence
                  Modeling},
  journal={arXiv preprint arXiv:1412.3555},
  year         = {2014}
}

@inproceedings{mpgsam,
  title={MPG-SAM 2: Adapting {SAM} 2 with mask priors and global context for referring video object segmentation},
  author={Fu Rong and
                  Meng Lan and
                  Qian Zhang and
                  Lefei Zhang},
  booktitle    = {Proceedings of IEEE/CVF International Conference on Computer Vision},
  pages={},
  year={2025}
}

@article{sa2va,
  author       = {Haobo Yuan and
                  Xiangtai Li and
                  Tao Zhang and
                  Zilong Huang and
                  Shilin Xu and
                  Shunping Ji and
                  Yunhai Tong and
                  Lu Qi and
                  Jiashi Feng and
                  Ming{-}Hsuan Yang},
  title        = {Sa2VA: Marrying {SAM2} with LLaVA for dense grounded understanding of images and videos},
  journal={arXiv preprint arXiv:2501.04001},
  year         = {2025}
}
